\documentclass[lettersize,journal]{IEEEtran}
\usepackage{amsmath,amsfonts}
\usepackage{mathtools}
\usepackage{textcomp}
\usepackage{stfloats}
\usepackage{url}
\usepackage{verbatim}
\usepackage{graphicx}
\usepackage{cite}
\usepackage{amsmath,amsfonts}
\usepackage{algorithmic}
\usepackage{algorithm}
\usepackage{array}
\usepackage[caption=false,font=footnotesize,labelfont=rm,textfont=rm]{subfig}
\usepackage{bbding}
\usepackage{booktabs}
\usepackage{multirow}
\usepackage{makecell}
\usepackage{amsmath,amsthm,amssymb,amsfonts}
\usepackage{threeparttable}
\usepackage{xcolor} 
\usepackage[colorlinks=true, linkcolor=blue, citecolor=blue, urlcolor=blue]{hyperref}

\begin{document}


\title{Prior-Driven Multi-Scale Causal Discovery in Topological Event Sequences for Anomaly Tracing}

\author{Jiyu Tian,~Mingchu Li,~Liming Chen,~\IEEEmembership{Senior Member,~IEEE},~M-Tahar Kechadi,~\IEEEmembership{Member,~IEEE}

\thanks{}

\thanks{Jiyu Tian and Mingchu Li are with the School of Software, Dalian University of Technology, Dalian, China. Mingchu Li is also affiliated with the School of Computer and Information Engineering, Jiangxi Normal University, Nanchang, Jiangxi, China. }

\thanks{Liming Chen is with the School of Computer Science and Technology, Dalian University of Technology, Dalian, China.}

\thanks{M-Tahar Kechadi is with the School of Computer Science,  University College Dublin, Dublin, Ireland.}
}
\markboth{ }
{How to Use the IEEEtran \LaTeX \ Templates}

\maketitle

\begin{abstract}
Identifying causal relationships in alarm events is critical for tracing anomalous behavior such as attacks and failures in telecommunications networks. However, existing causal discovery methods either fail to distinguish between long-term and short-term dependencies in anomaly propagation or incur massive computational overhead that cannot meet the real-time processing demands of large-scale alarm event sequences. To address this challenge, we propose a prior-driven multi-scale Granger causal discovery approach for topological event sequences, named PMCDTrace. Specifically, we first design a multi-scale Hawkes process with a piecewise constant approximation to model both short-term rapid propagation and long-term slow propagation. Furthermore, we develop a prior knowledge-driven long-short term constraint mechanism, which ensures the rationality and robustness of the causal structure through sparsity, acyclicity, and alignment constraints. To validate the effectiveness of PMCDTrace, we conduct experiments on 6 synthetic and 3 real-world telecommunication network alarm datasets. Experimental results demonstrate that our method outperforms state-of-the-art (SOTA) approaches in both causal discovery accuracy and computational efficiency. 

\end{abstract}

\begin{IEEEkeywords}
Security Forensics, Telecommunication Networks, Granger Causal Discovery
\end{IEEEkeywords}


\section{Introduction}
\label{sec:1}

\IEEEPARstart{I}{n} modern telecommunication networks, malicious activities and cascading failures typically trigger a massive number of alarms, and rapidly tracing anomalies is crucial for ensuring network security and stability \cite{threatrace, tered, tracecluster, tagapt, aptkcg}. Causal discovery can delineate relationships between alarm event types, forming a causal graph for abnormal behavior, which can assist operators in understanding anomaly propagation paths and thus efficiently identify root causes of attacks and failures \cite{cdsurveys1, cdsurvery2, llmcd}. As a representative causal discovery method, Granger causality analysis infers causal relationships between events by evaluating whether the historical information of one event type improves the predictability of another, serving as a crucial paradigm for learning causal structures from alarm event sequences \cite{atlas,flash,attackinves}.

\begin{figure}[h]
  \centering
  \includegraphics[width=0.96\linewidth]{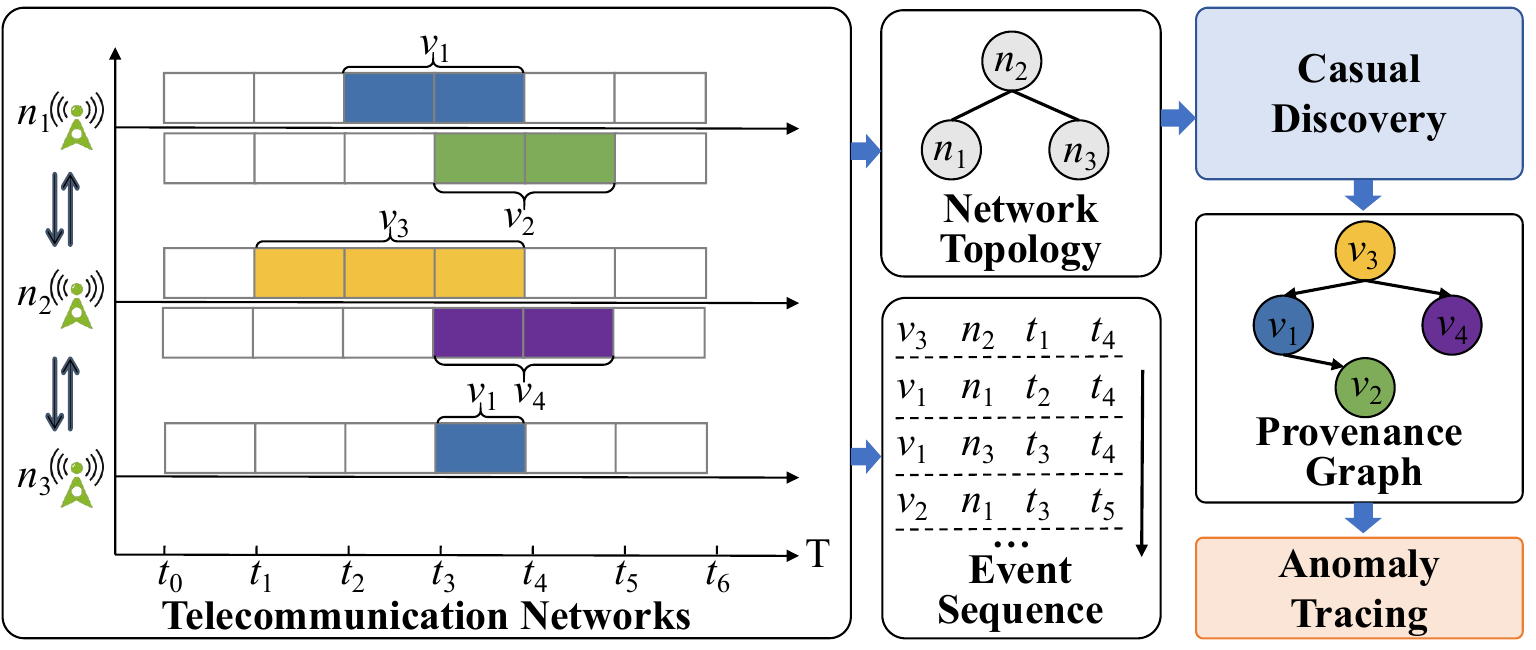}
  \caption{An illustrative example of causal discovery for alarm event sequences. Telecommunications networks consist of multiple device nodes ($n_1, n_2, n_3$) interconnected via physical links, where different alarm event types ($v_1, v_2, v_3, v_4$) occur over time on each device. The event sequences record the occurrence order and duration of these events along the timeline. By analyzing alarm sequences and network topology, the alarm event causal discovery method effectively reconstructs the source graph and ultimately infers causal relationships between alarm event types, thereby enabling anomaly tracing.}
  \label{figure1}
\end{figure}

However, as illustrated in Figure \ref{figure1}, telecommunication networks consist of numerous devices interconnected via physical links, forming a complex topological network. Meanwhile, different types of alarm events continuously occur over time on different devices, generating sequences with inherent dependencies. Consequently, the complexity of both topological structures and temporal data poses two major challenges for alarm event sequence causal discovery methods. First, devices are tightly coupled through the topology, leading to interdependencies among alarm event sequences that directly violate the independent and identically distributed (i.i.d.) assumption underpinning traditional approaches. However, anomaly propagation is strictly constrained by physical topology, necessitating that causal discovery explicitly incorporate topological network information. Second, anomaly propagation exhibits multi-scale characteristics, encompassing both rapid alarms that are instantly triggered between directly connected devices and delayed alarms that slowly propagate through multi-hop paths.

Existing methods have begun to incorporate topological information into causal learning \cite{thps, tnpar, casualnet, s2gcsl, tccd}. By leveraging neural point processes \cite{tnpar, casualnet} and Hawkes process modeling \cite{thps, s2gcsl, tccd}, they can characterize the occurrence intensity of temporal alarm events, partially addressing the issue of i.i.d. assumption violations. However, such methods still suffer from two key limitations. First, in terms of modeling capacity, existing approaches \cite{s2gcsl} either employ a linear Hawkes process, which fails to distinguish multi-scale characteristics such as short-term and long-term effects in anomaly propagation. Second, regarding computational efficiency, methods based on neural point processes \cite{tnpar, casualnet} and continuous-time Hawkes decay kernels \cite{thps, tccd} involve extremely complex likelihood calculations, rendering them unable to meet the efficiency demands of real-time diagnosis in large-scale telecommunication networks. We formalize these research challenges addressed in this work as follows:

\begin{itemize}
\item \textbf{Heterogeneity:} Existing methods lack multi-scale modeling capabilities when capturing differentiated anomaly propagation patterns. Specifically, while TNPAR \cite{tnpar} enhances model expressiveness using neural point processes, its intensity function lacks an explicit multi-scale decomposition mechanism, making it difficult to explicitly capture and separate short-term direct effects from long-term indirect effects in anomaly propagation. S$^2$GCSL \cite{s2gcsl} employs a linear kernel to simulate occurrence intensity among events, but its influence pattern is monolithic and fixed. Consequently, it cannot distinguish between instantaneous cascades triggered by quick scan attack and slow diffusion caused by persistent malicious activities, thereby compromising the accuracy of alarm event causal learning.

\item \textbf{Efficiency:} Existing methods face significant computational bottlenecks when processing large-scale topological alarm events. Specifically, THP \cite{thps} is based on a topological Hawkes process with a continuous-time decay kernel, requiring evaluation of all historical events to compute the intensity at any given moment, leading to substantial computational overhead. TCCD \cite{tccd} introduces a hybrid framework combining Monte Carlo tree search with gradient optimization alongside continuous-time exponential decay kernels. Although this enhances global exploration capabilities, its search process is computationally expensive, making it difficult to apply in telecommunication network security operations scenarios that demand high real-time performance.
\end{itemize}

To address the challenges above, we propose a \textbf{\underline{P}}rior-driven \textbf{\underline{M}}ulti-scale \textbf{\underline{C}}ausal \textbf{\underline{D}}iscovery approach for topological event sequences for anomaly \textbf{\underline{T}}racing  (\textbf{PMCDTrace}). Specifically, we design a multi-scale Hawkes process (MSHP) to replace traditional linear kernels or continuous decay kernels. By adopting a piecewise constant approximation, historical events are partitioned into multiple scales based on their time delays, with independent causal matrices and learnable weights assigned to each scale to differentiate between short-term rapid propagation and long-term slow accumulation of malicious infiltration and cascading failures. In addition, we introduce a long-short term constraint (LSTC) mechanism that incorporates regularization terms for sparsity, acyclicity, and alignment constraints, thereby ensuring that the learned provenance graph exhibits sparsity, logical consistency, and cross-scale robustness. Experimental results demonstrate that our method achieves superior accuracy and efficiency in causal discovery for abnormal provenance analysis when handling complex topological time series. In summary, the main contributions of this paper are as follows:

\begin{itemize}
\item We propose a multi-scale Hawkes process that models both short-term and long-term effects in anomaly propagation through piecewise constant approximation and scale-independent causal matrices.

\item We design a long-short term constraint mechanism that enforces sparsity, acyclicity, and alignment regularization based on prior knowledge, ensuring the physical interpretability of the learned multi-scale causal graph.

\item We conduct comprehensive experiments on 6 synthetic datasets and 3 real-world telecommunication network alarm datasets, validating that the proposed method outperforms existing baselines in both causal discovery accuracy and computational efficiency.

\end{itemize}

The remainder of this paper is organized as follows. Section~\ref{sec:2} reviews related work. Section~\ref{sec:3} outlines the overall framework of the PMCDTrace method. Section~\ref{sec:4} theoretically analyzes the effectiveness of PMCDTrace. Section~\ref{sec:5} presents the evaluation results of the proposed method. Section~\ref{sec:6} discusses the effectiveness and limitations of the method. Finally, Section~\ref{sec:7} concludes the paper with a summary of our contributions.

\section{Related Work}
\label{sec:2}
\subsection{Temporal Point Process}
\label{sec:2.1}
Temporal point processes are fundamental tools for characterizing the occurrence intensity of discrete event sequences, transforming causal discovery into an optimizable problem \cite{tppsurvey1, tppsurveys2, tppsurveys3, ithpp}. Currently, based on modeling paradigms, temporal point processes can be categorized into statistical point processes \cite{tittpp, mhpg, sehp} and neural point processes \cite{unipoint, dtpp, teise}.

Statistical point processes focus on constructing intensity functions with explicit physical interpretations. Among these, the Poisson process \cite{poisson} assumes that events occur randomly at a constant rate independent of historical events, serving as the foundational model for point processes. Reactive point process \cite{rpps} simulates complex excitatory and inhibitory relationships among events through self-exciting, self-regulating, and saturation mechanisms to predict discrete events. Hawkes process \cite{hp} extends the Poisson process by allowing historical events to excite future occurrences, making it central to causal modeling. Zhang et al. \cite{npbhp} method further proposes a non-parametric Bayesian inference method for Hawkes processes, transforming it into a cluster of Poisson processes for efficient inference via branching structure sampling. ELBO \cite{elbo} method formulates the triggering kernel as a sparse Gaussian process and leverages variational inference to achieve optimization with linear time complexity. MML \cite{mml} method describes multivariate Hawkes processes with exponential decay kernels using connectivity graphs and provides a model selection algorithm for estimating Granger causality based on the minimum message length principle.

Neural point processes utilize neural networks to parameterize intensity functions, liberating them from reliance on pre-defined kernel functions. RMTPP \cite{rmtpp} method pioneers the use of recurrent neural networks to model event histories, treating the intensity function as a non-linear mapping of historical information. To simultaneously capture both long-term and short-term dependencies, THP \cite{thp} method introduces a self-attention mechanism and proposes the transformer-based Hawkes process method (Note that THP and the subsequent THPs are not the same method).  SAHP \cite{sahp} method further improves positional encoding by incorporating event time intervals into attention computation, enhancing model interpretability. CAUSE \cite{cause} method adopts a two-stage framework that first trains a neural point process model and then extracts causal statistics based on attribution analysis. PromptTPP \cite{prompttpp} method innovatively combines large language models with continual learning, employing a learnable prompt pool for continuous learning from streaming events. RUN \cite{run} method integrates contextual information from time series, leverages contrastive learning to enhance the encoder, and performs neural Granger causal discovery through prediction tasks.

However, existing statistical point process methods either rely on linear kernels that neglect short-term and long-term effects of alarm events or employ complex kernel functions such as exponential decay, making it difficult to ensure causal discovery efficiency in large-scale telecommunication networks. Meanwhile, the computational complexity inherent in neural point process models severely limits their scalability when processing massive alarm sequences from telecommunication networks.

\subsection{ Casual Discovery for Alarm Events}
\label{sec:2.2}
Causal discovery in event sequences can reveal causal relationships between alarm events, which is crucial for anomaly tracing \cite{lcea, mdl, tkdecd, causalformer, lshcd, llmgc}. Based on whether network topology information is incorporated, existing causal discovery methods for event sequences can be categorized into two classes: causal discovery without topology information \cite{busca, l0hawkes, cutsplus} and causal discovery with topology information \cite{anhp}.

Casual discovery without topology information typically assume that event sequences are independent and identically distributed. PC \cite{pc} algorithm constructs causal graphs through conditional independence tests and searches for the optimal directed acyclic graph based on Bayesian Information Criterion and Minimum Description Length scores. PCMCI+ algorithm \cite{pcmci} improves the selection of conditioning sets, enhancing the efficiency and reliability of causal discovery in highly autocorrelated time series. GraNDAG \cite{grandag} method formulates graph learning as a continuous optimization problem, seeking the optimal structure by maximizing a score function, albeit with high computational cost. Multivariate Hawkes processes based on Granger causality are widely applied, yet they struggle to handle instantaneous effects caused by low-resolution data in discrete-time settings. SHPs \cite{shps} method addresses causal identifiability issues in discrete Hawkes processes by explicitly modeling instantaneous effects. CASCADE \cite{cascade} method identifies causal network structures based on the principle of minimum description length by minimizing the complexity of data descriptions.

Casual discovery with topology information incorporates the known topological network as prior knowledge into the model. THPs \cite{thps} method first integrates network topology into causal discovery by introducing graph convolutions into the Hawkes process, forcing event influences to propagate along topological edges, and employs a gradient-free optimization strategy to search for causal structures, laying the foundation for modeling topological event sequences. Building upon THPs, the TNPAR \cite{tnpar} method introduces neural point processes to enhance modeling capacity and treats the causal structure as a latent variable, enabling unified modeling of both the topological network and the underlying causal graph. CauseNET \cite{casualnet} method constructs a Transformer-based prediction module, utilizing a trainable causal graph and decay matrices encapsulating topological influences, and learned through an alternating optimization strategy. S$^2$GCSL \cite{s2gcsl} method employs a linear kernel with gradient descent to efficiently optimize the likelihood function, achieving state-of-the-art performance in both efficiency and effectiveness. To address non-stationarity, StateHPs \cite{statehps} method models sequence states as Dirichlet distributions and assigns distinct Hawkes processes to each state, identifying causality via a variational EM algorithm. To overcome the issue of exploding search space, TCCD \cite{tccd} method integrates Monte Carlo tree search with continuous gradient optimization, balancing global exploration and local refinement.

However, although existing methods have successfully incorporated topological constraints, they either suffer from inefficiencies in their intensity functions, model structures, and optimization strategies, or fail to model and differentiate between long-term and short-term anomaly propagation patterns in telecommunication networks along the temporal dimension.

\section{Methodology}
\label{sec:3}

\subsection{Problem Definition}
\label{sec:3.1}
Given a telecommunication network, its topological structure is represented as an undirected graph $\mathcal{G}_N = (\mathcal{N}, \mathcal{E}_N)$, where $\mathcal{N}$ denotes the set of devices and $\mathcal{E}_N$ represents the physical connections between them. Let $\mathcal{V}$ be the set of event types. Each occurred event is represented as a tuple $(v_i, n_i, t_i, \gamma_i)$, where $v_i \in \mathcal{V}$ is the event type, $n_i \in \mathcal{N}$ is the device on which the event occurs, $t_i \in [0,T]$ is the start timestamp, and $\gamma_i \in \mathbb{Z}^+$ denotes the duration of an event $v_i$. The observed event sequence is denoted as $\mathbf{X} = {(v_i, n_i, t_i, \gamma_i) \mid i = 1,\dots,m}$. The topological adjacency matrix $\mathbf{B}^k \in \{0,1\}^{|\mathcal{N}| \times |\mathcal{N}|}$ characterizes the $k$-hop neighbor relationships between devices, where $\mathbf{B}^k_{n_i,n_j}=1$ if device $n_j$ is within the $k$-hop range of device $n_i$. The multi-scale Granger causal matrix is denoted as $\mathbf{A}^m \in \mathbb{R}^{|\mathcal{V}|\times|\mathcal{V}|}$. 

Our objective is to learn a causal matrix $\mathbf{A}$ that optimally explains the causal relationships among different alarm event types in $\mathcal{V}$, based on the event sequences $\mathbf{X}$ and under the constraints imposed by the network topology $\mathbf{B}^k$.

\begin{figure*}
\centering
\includegraphics[width=0.98\linewidth]{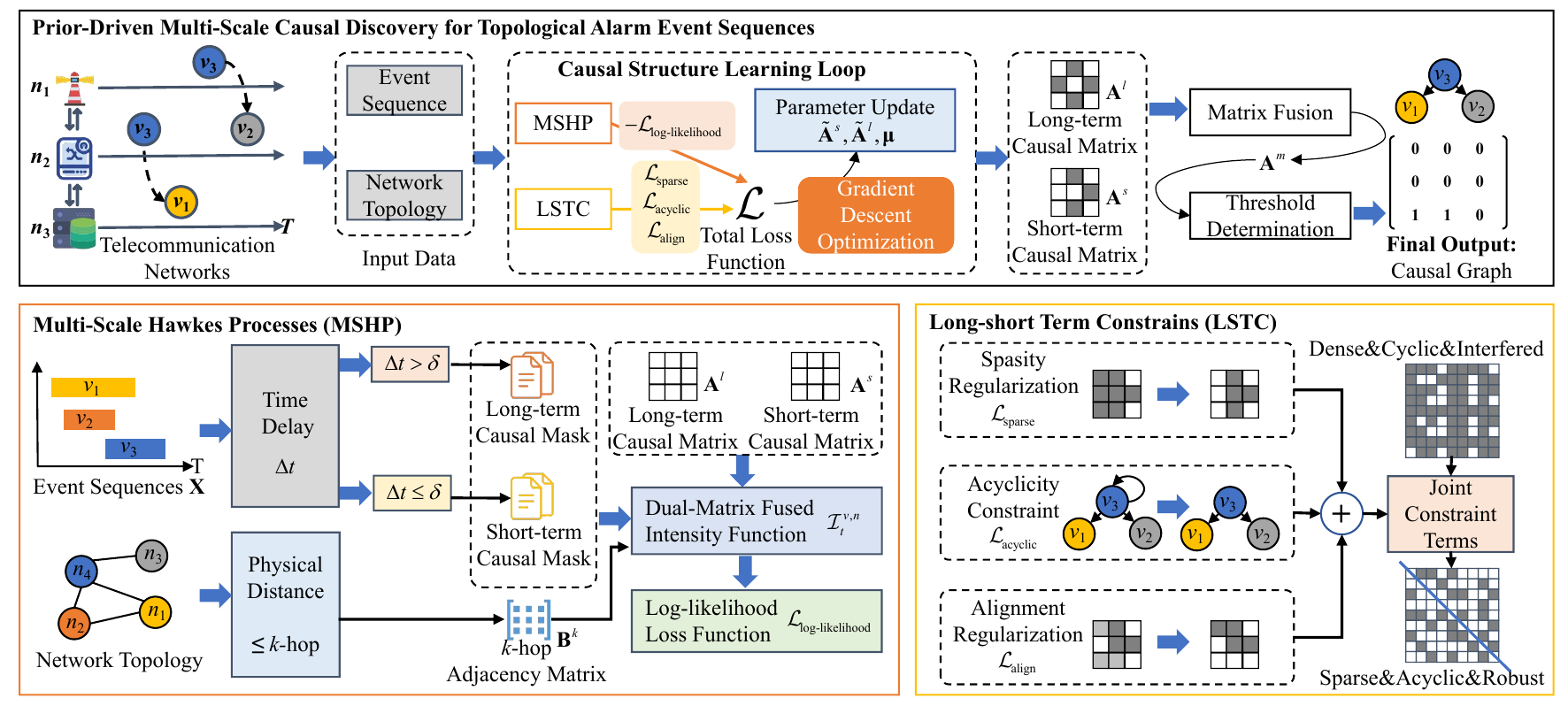}
\caption{Overview of PMCDTrace approach, which composed of two main components: the multi-scale Hawkes process, which partitions historical events into different scales based on time delays via piecewise constant approximation and assigns independent causal matrices as learnable weights to each scale; and the long-short term constraint mechanism, which imposes sparsity, acyclicity, and alignment regularizations to ensure that the learned causal relationships exhibit sparsity, acyclicity, and cross-scale consistency.} 
\label{framework}
\end{figure*}

\subsection{Overview}
\label{sec:3.2}
In this paper, we propose a prior-driven multi-scale Granger causal learning approach for topological alarm event sequences, which jointly addresses the performance and efficiency challenges in causal discovery for abnormal provenance analysis. As illustrated in Figure \ref{framework}, our method first models cross-device and cross-type causal relationships among events through a multi-scale Hawkes process. Subsequently, a prior knowledge-driven long-short term constraint mechanism is introduced to refine the structure of the causal graph, ultimately yielding a sparse and acyclic event causal graph. The proposed method comprises two novel components: the Multi-scale Hawkes Process (MSHP) and the Long-Short Term Constraints (LSTC).

\textbf{MSHP:} To capture both the instantaneous rapid anomaly propagation and long-term persistent malicious infiltration, we partition historical events into two scales based on time delays: short-term ($\Delta t \leq \delta$) and long-term ($\Delta t > \delta$).  In practice, topological distance constraints and multi-scale time are realized by two causal masks. Furthermore, we maintain two distinct causal matrices, $\mathbf{A}^s$ and $\mathbf{A}^l$, for each scale, respectively. A dual-matrix fused Hawkes process is then designed as the intensity function, enabling the model to adaptively capture both rapidly propagating local anomalies and slowly diffusing systemic anomalies.

\textbf{LSTC:} To ensure that the learned short-term and long-term causal graphs align with domain prior knowledge, we integrate three regularization terms with the likelihood function as the loss for optimization. Specifically, sparsity regularization encourages each event type to influence only a few others; acyclicity regularization enforces the causal graph to be a directed acyclic graph; and alignment regularization encourages consistency in causal patterns across different scales, mitigating conflicts caused by noise. The final objective function is the sum of the negative log-likelihood and the regularization terms, which is optimized end-to-end via gradient descent.

Overall, the raw event sequence, combined with the topological network, yields the $k$-hop adjacency matrix $\mathbf{B}^k$ between devices. Next, MSHP assigns the historical context of each event to different scales based on time delays. Then, the intensity function computes the log-likelihood contribution of each event using the current parameters, contributing to the overall loss. Subsequently, the log-likelihood loss is combined with the regularization term from LSTC to update the causal matrix and baseline intensity parameters. Finally, through threshold-based fusion of the multi-scale causal matrices, the final causal graph is obtained as the output.

\subsection{Multi-scale Hawkes Process}
\label{sec:3.3}

In telecommunication networks, anomaly propagation exhibits complex temporal characteristics: some abnormal effects are instantaneous, while others may manifest after a certain period. Consequently, unlike temporal point processes that rely on predicting event occurrence probabilities or linear Hawkes processes with simple predictive mechanisms, we adopt a different approach. For a current event $v_i$, we consider all neighbor events $v_j$ that are active at time $t_i$, i.e., those satisfying $t_j < t_i \leq t_j + \gamma_j$ and $\mathbf{B}^k_{n_i,n_j}=1$.

To distinguish influences across different time scales, we partition historical events into two sets based on the time difference $\Delta t = t_i - t_j$:

\begin{itemize}
    \item \textbf{Short-term influence event set:} $\mathbf{A}^s(t_i) = \{j \mid t_j < t_i \leq t_j + \gamma_j, \Delta t \leq \delta\}$, modeling rapid direct anomaly propagation.
    \item \textbf{Long-term influence event set:} $\mathbf{A}^l(t_i) = \{j \mid t_j < t_i \leq t_j + \gamma_j, \Delta t > \delta\}$, modeling slow and persistent abnormal effects.
\end{itemize}

Here, $\tau$ is a weight parameter balancing the importance of the short-term and long-term matrices. Accordingly, we define the intensity function for event type $v$ on device $n$ at time $t$ as:

\begin{equation}
\begin{split}
\mathcal{I}_{t}^{v,n} = \mu^{v,n} + \sum_{\substack{i:t_i<t\le t_i+\gamma_i \\[1pt]}} \alpha_{v_i v} \Bigl[ & \tau \mathbf{A}_{v_i,v}^{s} \mathbb{I}_{\{t-t_i<\mathcal{T}\}} \\
& \hspace*{-4em} + (1-\tau) \mathbf{A}_{v_i,v}^{l} \mathbb{I}_{\{t-t_i\ge\mathcal{T}\}} \Bigr] \mathbf{B}_{n_i,n}^{k}
\end{split}
\end{equation}

where $\mu^{v,n}$ is the baseline intensity, representing the inherent likelihood of such a abnormal occurring on the device itself even without any external influence. $\mathbf{A}^s$ and $\mathbf{A}^l$ are the corresponding short-term and long-term causal matrices, respectively, and $\tau$ is a weight parameter balancing the importance of the short-term and long-term matrices, $\tau \in [0,1]$. The key innovation lies in the definition of the intensity, which no longer relies on a single or continuously decaying excitation kernel. Instead, it is composed of the baseline intensity together with a multi-scale excitation term. This excitation term employs two independent causal matrices, along with the temporal indicator functions, to separately characterize the rapid, direct occurrence intensity from recent historical events and the slow, indirect, or persistent influences from distant historical events. The term $\mathbf{B}_{n_i,n}^{k}$ ensures that influences propagate only along physical network connections (within $k$ hops). In practice, we typically assign higher weights to short-term events, aligning with the industrial intuition that recent influences are more significant.

To derive a computationally tractable optimization objective for learning causal relationships from data, we design a multi-scale log-likelihood function. Given an event sequence $\mathbf{X}$, maximizing this likelihood function is equivalent to finding an optimal set of parameters:

\begin{equation}
\mathcal{S}_{i}^{v_i,n} = \sum_{\mathclap{j: t_j<t_i\le t_j+\gamma_j}} \bigl( \tilde{\mathbf{A}}_{v_i v}^{s} \mathbb{I}_{\{t-t_i<\mathcal T\}} + \tilde{\mathbf{A}}_{v_i v}^{l} \mathbb{I}_{\{t-t_i>\mathcal T\}} \bigr) \mathbf{B}_{n_j n_i}^{k}
\end{equation}

\begin{align}
\mathcal{R}_{i}^{v,n} = 
\Bigl( \tilde{\mathbf{A}}_{v_i,v}^{s} \mathbb{I}_{\{t-t_i<\mathcal{T}\}} 
+ \tilde{\mathbf{A}}_{v_i,v}^{l} \mathbb{I}_{\{t-t_i>\mathcal{T}\}} \Bigr) \mathbf{B}_{n_i,n}^{k} \gamma_i
\end{align}

\begin{flalign}
&\mathcal{L}_{\text{log-likelihood}}(\tilde{\mathbf{A}}^{s},\tilde{\mathbf{A}}^{l},\boldsymbol{\mu}) \nonumber \\
&\quad\quad\quad\quad =\sum_{n}\left(\sum_{i:n_i=n}\log\mathcal{I}_{t_i}^{v_i,n}-\sum_{v}\int_{0}^{T}\mathcal{I}_{t}^{v,n}dt\right) \nonumber\\
&\quad\quad\quad\quad=\sum_n \Biggl( \sum_{i:n_i=n} \log\bigl( \mu^{v_i,n} + \mathcal{S}_{i}^{v_i,n} \bigr) \nonumber - T \sum_v \mu^{v,n} \\
&\quad\quad\quad\quad\quad 
- \sum_v \sum_{i:n_i=n} \mathcal{R}_{i}^{v,n} \Biggr)
\label{formula4}
\end{flalign}

where $\Theta = \{\mu, \mathbf{A}^s, \mathbf{A}^l\}$ denotes the parameters to be optimized. In formula \ref{formula4}, the first term $\sum_{i:n_i=n}\log\mathcal{I}_{t_i}^{v_i,n}$ sums over all events that occur, rewarding the model for accurately predicting the observed sequence of events. If the model accurately predicts these actual events, the value of this term will be large. The second term $\sum_{v}\int_{0}^{T}\mathcal{I}_{t}^{v,n}dt$ is the integral of the intensity function over the time interval $[0, T]$, calculating the total number of anomalies predicted by the model across the entire period. This term penalizes over-prediction, preventing the model from simply inflating all intensities to fit the data, thereby encouraging the learning of precise and necessary causal connections while avoiding redundant associations.

The optimization of the log-likelihood loss function seeks an optimal balance between rewarding the model for accurately reproducing observed data and penalizing it for making excessive false predictions. By designing a multi-scale intensity function and maximizing its corresponding likelihood function, we enable the model to distinguish between fast and slow propagation modes.

\subsection{Long-Short Term Constraints}
\label{sec:3.4}
The proposed MSHP enhances the predictive capability of event point processes by capturing both fast and slow propagation modes. However, without constraints, the model is prone to learning numerous spurious causal relationships arising from noise and coincidences in complex telecommunication network alarm data. This can lead to an excessively dense causal graph that violates practical logical directionality and exhibits inconsistencies between short-term and long-term causality. Therefore, to obtain a causality structure with physical significance, we introduce three domain-specific prior knowledge elements: sparsity, directionality, and cross-scale consistency. Simultaneously, we propose three corresponding long-term and short-term regularization constraints.

In telecommunication networks, an abnormal behavior typically triggers only a few related alarms rather than affecting all event types. Consequently, the causal graph should be sparse. Following the work on single-scale sparsity constraints \cite{s2gcsl}, we simultaneously enforce multi-scale sparsity on both short-term and long-term causal matrices through weighted L$1$-norm constraints:

\begin{equation}
\mathcal{L}_{\text{sparse}}=\lambda\sum_{i=1}^{|\mathcal{V}|}\sum_{j=1}^{|\mathcal{V}|}\left|\tau\cdot\tilde{\mathbf{A}}_{i,j}^{s}+(1-\tau)\cdot\tilde{\mathbf{A}}_{i,j}^{l}\right|
\end{equation}

where $\lambda$ controls the sparsity strength. The sparsity regularization term constrains the weighted sum of the two matrices rather than applying the L$1$-norm to each separately, encouraging them to share a similar sparse pattern overall, aligning with the domain knowledge that anomaly propagation has limited influence. Given that causal relationships should exhibit a clear temporal direction without forming cycles, anomaly propagation in telecommunication networks should constitute a directed acyclic graph (DAG). To ensure the causal graph is a DAG, we further refine the single-scale acyclicity constraints from \cite{s2gcsl} into multi-scale acyclicity constraints to:

\begin{equation}
\mathcal{L}_{\text{acyclic}}=\operatorname{trace}\left[(\mathbf{I}+\beta\tilde{\mathbf{A}}\circ\tilde{\mathbf{A}})^{|\mathcal{V}|}\right]-|\mathcal{V}|
\end{equation}

Here, $\mathbf{I}$ is the identity matrix, $\beta>0$ is a scaling factor, and $\circ$ denotes element-wise multiplication. $\mathcal{L}_{\text{acyclic}}$ loss function equals zero if and only if the fused causal matrix $\tilde{\mathbf{A}}$ corresponds to a DAG. Since the same abnormal behavior may generate both short-term and long-term effects, the causal structures at the two time scales should exhibit a certain degree of consistency. The alignment regularization encourages the short-term and long-term matrices to remain consistent in their core causal relationships:

\begin{equation}
\mathcal{L}_{\text{align}}=\sum_{i=1}^{|\mathcal{V}|}\sum_{j=1}^{|\mathcal{V}|}\left|\tilde{\mathbf{A}}_{i,j}^{s}-\tilde{\mathbf{A}}_{i,j}^{l}\right|
\end{equation}

where the regularization term $\mathcal{L}_{\text{align}}$ computes the L$1$-norm of the element-wise differences between the two matrices. Minimizing this term encourages the short-term and long-term matrices to be as similar as possible, preventing the model from learning contradictory causal patterns from noisy data and thus enhancing the robustness of the results. Combining the negative log-likelihood with the regularization terms, we obtain the final loss function as follows:

\begin{align}
\mathcal{L}=&-\mathcal{L}_{\text{log-likelihood}}+\mathcal{L}_{\text{sparse}}+\mathcal{L}_{\text{acyclic}}+\mathcal{L}_{\text{align}} \nonumber\\
\text{s.t.}\quad &\mathcal{L}_{\text{sparse}}, \mathcal{L}_{\text{align}} \le \epsilon,\mathcal{L}_{\text{acyclic}}=0
\end{align}

The loss function comprises three regularization terms. Specifically, we require the sparsity regularization term not to exceed a pre-specified small positive constant $\epsilon$, thereby ensuring the sparsity of the causal graph by limiting the number of non-zero connections. The alignment regularization ensures robustness by enforcing consistency between the short-term and long-term matrices. Meanwhile, we strictly enforce the acyclicity regularization term to be zero, guaranteeing the acyclicity of the final inferred causal graph. After optimizing the causal matrices at different scales, we obtain a unified causal matrix via weighted fusion:

\begin{equation}
\mathbf{A}^{m}=\tau\mathbf{A}^{s}+(1-\tau)\mathbf{A}^{l}
\end{equation}

After fusing the long-term and short-term causal matrices, we finally binarize the continuous-valued matrix using a threshold $\rho$:

\begin{equation}
\mathbf{A}=(a_{i,j}),\quad\text{where }a_{i,j}= \begin{cases}
1 & \text{if }\mathbf{A}_{i,j}^{m}>\rho,\\
0 & \text{if }\mathbf{A}_{i,j}^{m}\le\rho
\end{cases}
\end{equation}

Here, $\mathbf{A}$ denotes the inferred Granger causal graph adjacency matrix. The threshold $\rho$ is a pre-defined positive hyperparameter. Connections with intensity scores above $\rho$ are considered to have significant causal relationships between event types and are retained in the graph (value 1); connections with scores below $\rho$ are considered to have no causal relationship.

\section{Theoretical Analysis}
\label{sec:4}
Under the modeling framework of Granger causal discovery from topologically constrained alarm event sequences, our proposed multi-scale Hawkes process achieves strictly stronger expressive power than the linear Hawkes process (LHP) while introducing negligible additional computational overhead. We provide theoretical proofs regarding the performance and efficiency of MSHP as follows.

\textbf{Proposition.} Let $\mathcal{H}_L$ and $\mathcal{H}_M$ denote the model spaces of the linear Hawkes process and the multi-scale Hawkes process, respectively. For a telecommunication network event sequence generated by an intensity function $\mathcal{I}^*$ that inherently exhibits multi-scale propagation characteristics, if the optimal Kullback–Leibler (KL) approximation of $\mathcal{I}^*$ within $\mathcal{H}_M$, defined as $\mathcal{I}_M := \arg\min_{\mathcal{I} \in \mathcal{H}_M} D_{\text{KL}}(\mathcal{I}^* \| \mathcal{I})$, satisfies $\mathbf{A}^{s*} \neq \mathbf{A}^{l*}$, then MSHP yields a strictly smaller optimal KL approximation error. That is,
\[
\inf_{\mathcal{I} \in \mathcal{H}_M} D_{\text{KL}}(\mathcal{I}^* \| \mathcal{I}) \;<\; \inf_{\mathcal{I} \in \mathcal{H}_L} D_{\text{KL}}(\mathcal{I}^* \| \mathcal{I}).
\]

\textbf{Proof.} First, note that when we impose the constraint $\mathbf{A}^s = \mathbf{A}^l$ on the parameters of MSHP, its intensity function degenerates exactly to that of LHP. Hence, the model space of LHP is a subset of that of MSHP, i.e., $\mathcal{H}_L \subset \mathcal{H}_M$. By the definition of the KL divergence, the infimum of the divergence over a larger space $\mathcal{H}_M$ cannot exceed that over its subspace $\mathcal{H}_L$. Consequently, the set inclusion directly implies the inequality
\[
\inf_{\mathcal{I} \in \mathcal{H}_M} D_{\text{KL}}(\mathcal{I}^* \| \mathcal{I}) \leq \inf_{\mathcal{I} \in \mathcal{H}_L} D_{\text{KL}}(\mathcal{I}^* \| \mathcal{I}).
\]

According to the premise, the real anomaly propagation process in telecommunication networks exhibits distinct characteristics such as instantaneous cascades and slow diffusion. This implies that its optimal approximation $\mathcal{I}_M$ within the full space $\mathcal{H}_M$ must rely on different scale matrices (i.e., $\mathbf{A}^{s*} \neq \mathbf{A}^{l*}$), and therefore $\mathcal{I}_M \notin \mathcal{H}_L$. If the above inequality were to hold with equality, it would mean that there exists a model in the subspace $\mathcal{H}_L$ achieving the same global minimum KL divergence as $\mathcal{I}_M$. This contradicts the fact that $\mathcal{I}_M$ is the unique best approximation point in $\mathcal{H}_M$ and lies outside $\mathcal{H}_L$. Hence, the inequality is strict. In summary, because MSHP possesses a richer hypothesis space, it can approximate the true data-generating process with multi-scale characteristics more closely than LHP.

\textbf{Proposition.} Let the observed event sequence be $\mathbf{X}$, consisting of $M$ events. Let the average sparsity of the topological adjacency matrix $\mathbf{B}^k$ be $d$ (i.e., each device has on average $d$ $k$-hop neighbors). In a single iteration, computing the negative log-likelihood $\mathcal{L}$ of the MSHP model and its gradients with respect to the parameters $\Theta = \{\boldsymbol{\mu}, \mathbf{A}^s, \mathbf{A}^l\}$ has a time complexity of $O(d \cdot M)$.

\textbf{Proof. }For each event $i$, we need to aggregate the influence from its historical events $j$. Due to the topological constraint $\mathbf{B}^k$, each event only needs to examine historical events occurring on an average of $d$ neighboring devices. Determining whether an event falls into the short-term or long-term set is an $O(1)$ operation. Thus, computing $\mathcal{I}_{t_i}$ for all $M$ events incurs a complexity of $O(d \cdot M)$. After all intensities are computed, the evaluation of the log-likelihood $\mathcal{L}_{\text{log-likelihood}}$ and its gradients can be performed in $O(d \cdot M + |\mathcal{N}| \cdot |\mathcal{V}|)$ time. It is important to note that in typical telecommunication network scenarios, $M \gg |\mathcal{N}| \cdot |\mathcal{V}|$; hence, the dominant term is $O(d \cdot M)$, which is of the same order as that of the LHP. Therefore, while introducing multi-scale expressiveness, MSHP maintains computational complexity comparable to that of the linear model.

\section{Experiment}
\label{sec:5}
In this section, we evaluate the proposed method by addressing the following five research questions:

RQ1: How does the proposed method perform in causal discovery on synthetic alarm datasets?

RQ2: How does the proposed method perform in causal discovery on real-world alarm datasets?

RQ3: What is the computational efficiency of the proposed method?

RQ4: How sensitive is the proposed method to hyperparameters?

RQ5: What is the contribution of each component to PMCDTrace?

\subsection{Dataset}
\textbf{Synthetic Datasets: }To evaluate the performance of our method, we generate synthetic telecommunication network alarm event sequences using the widely adopted gCastle API~\cite{gcastle}. gCastle allows us to control the scale of telecommunication devices, the number of event types, the strength of causal influences, and the volume of data, thereby examining the scalability and robustness of our approach. The data generation follows the procedure below: first, a Granger causal graph and a device topology network are randomly generated; second, root event records are generated using a Poisson process based on the baseline intensity parameters of the Hawkes process; finally, propagated event records are generated discretely according to the time interval $\Delta$ and the excitation strength $\alpha$. To approximate the sparsity of alarm event sequences in real-world scenarios, we introduce the parameter $\Delta$ to discretize the continuous time domain and aggregate events within the same timestamp. Following the works of~\cite{tnpar,s2gcsl,tccd}, and TCCD~\cite{tccd}, we adopt the default parameter settings: the number of event types is $20$, the number of devices is $40$, the baseline intensity $\mu \in [3\times 10^{-5}, 5\times 10^{-5}]$, the excitation strength $\alpha \in [0.02, 0.03]$, the generation period $T = 2$ days, and the discretization interval $\Delta = 2$. All experiments on synthetic datasets are conducted under the principle of controlled variables, adjusting only one parameter at a time while keeping the others at their default values.

\textbf{Metropolitan Cellular Network Alarm Dataset: }To validate the generalization ability of our method in real-world scenarios, we employ alarm data collected from operational metropolitan telecommunication networks \cite{s2gcsl}. Specifically, we use three datasets: \textbf{18V-55N-Wireless}: Contains 18 event types, 55 network devices, and 34,839 events. \textbf{24V-439N-Microwave}: Contains 24 event types, 439 network devices, and 64,599 events. \textbf{25V-474N-Microwave}: Contains 25 event types, 474 network devices, and 48,573 events. All datasets are generated based on real topological networks and underlying causal structures. Due to the logging characteristics of network devices, the event timestamps exhibit specific discretization intervals. These datasets differ in scale, network density, and event sparsity, enabling a comprehensive evaluation of causal discovery performance in complex real-world environments.

\subsection{Baselines and Evaluation Criteria}
To demonstrate the effectiveness of the proposed method, we extensively compare PMCDTrace with seven baseline methods: \textbf{PC}~\cite{pc}: A classic causal discovery algorithm based on conditional independence tests, which constructs a causal graph by progressively removing edges. \textbf{GraNDAG}~\cite{grandag}: A neural network-based continuous optimization method that effectively handles non-linear relationships among variables. \textbf{CORL}~\cite{corl}: A reinforcement learning-based causal structure search method that incrementally builds the causal graph through a reward mechanism. \textbf{THPs}~\cite{thps}: The first method to integrate topological networks into the Hawkes process, modeling event propagation via graph convolutions.
\textbf{TNPAR}~\cite{tnpar}: Builds upon THP by introducing neural point processes to enhance expressive capacity. \textbf{S$^2$GCSL}~\cite{s2gcsl}: A lightweight topological causal learning approach that employs a linear kernel and gradient descent. \textbf{TCCD}~\cite{tccd}: Combines Monte Carlo tree search with continuous gradient optimization, balancing global exploration and local refinement.

To evaluate the performance of causal discovery methods, we adopt four core evaluation metrics: Precision (PRE), Recall (REC), F1 Score (F1), and Structural Hamming Distance (SHD). In the binarized prediction of the causal graph, we define: 
\begin{itemize}
    \item TP: Correctly predicted causal edges.
    \item FN: Incorrectly predicted causal edges.
    \item TN: Correctly identified absent edges.
    \item FP: Incorrectly predicted absent edges.
\end{itemize}
\textbf{Precision} measures the accuracy of predicted edges, calculated as $\text{TP} / (\text{TP} + \text{FP})$. \textbf{Recall} measures the coverage of true edges, calculated as $\text{TP} / (\text{TP} + \text{FN})$. The \textbf{F1 Score} is the harmonic mean of precision and recall, providing a balanced evaluation of prediction performance. \textbf{SHD} quantifies the minimum number of edge operations (insertions, deletions, or reversals) required to transform the predicted graph into the true graph, directly reflecting overall structural discrepancy—lower values indicate better performance. These metrics collectively assess causal discovery results from the perspectives of prediction accuracy and structural integrity.

\subsection{Implementation and Environment}
We conducted all experiments on a workstation equipped with an Intel\textsuperscript{\textregistered} Xeon\textsuperscript{\textregistered} Gold 6326 CPU, 503 GB RAM, and NVIDIA L40 GPUs*8. The software environment consisted of Ubuntu 22.04, CUDA 12.6, Python 3.9.21, and PyTorch 2.6.0. During model training, we set the hyperparameters as follows: the number of topological neighbor hops $k = 2$, the time threshold for partitioning multi-scale influences $\delta = 200$, and the fusion weight for the short-term and long-term causal matrices $\tau = 0.8$. The regularization weights for sparsity, acyclicity, and alignment were all set to $5 \times 10^3$, and the baseline intensity $\boldsymbol{\mu}$ was initialized to $0.001$. We optimized the model using the Adam optimizer with a learning rate of $0.005$ and trained for $3000$ iterations to ensure sufficient convergence. For each dataset, we use sklearn’s precision-recall curve function to calculate precision-recall pairs at different thresholds $\rho$, and select the threshold that maximizes the validation set’s F1 score. 

\subsection{RQ1: Performance Comparison on Synthetic Datasets}

\begin{figure*}[!htb]
	\centering
	\scriptsize
	\subfloat[Sensitivity to Event Types (F1 Score)]{
		\includegraphics[width=0.32\linewidth]{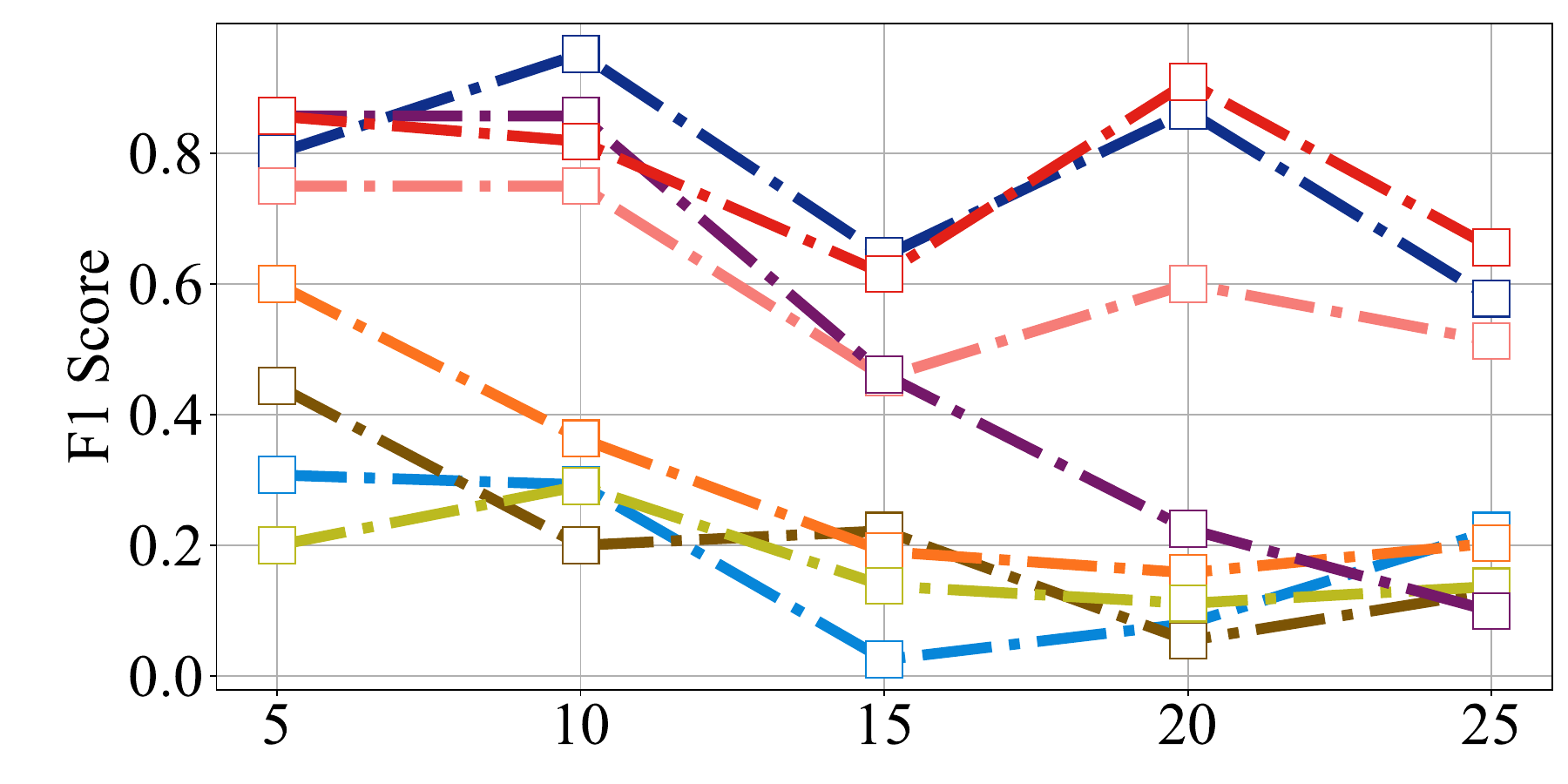}
	}
	\hfill
	\subfloat[Sensitivity to Devices (F1 Score)]{
		\includegraphics[width=0.32\linewidth]{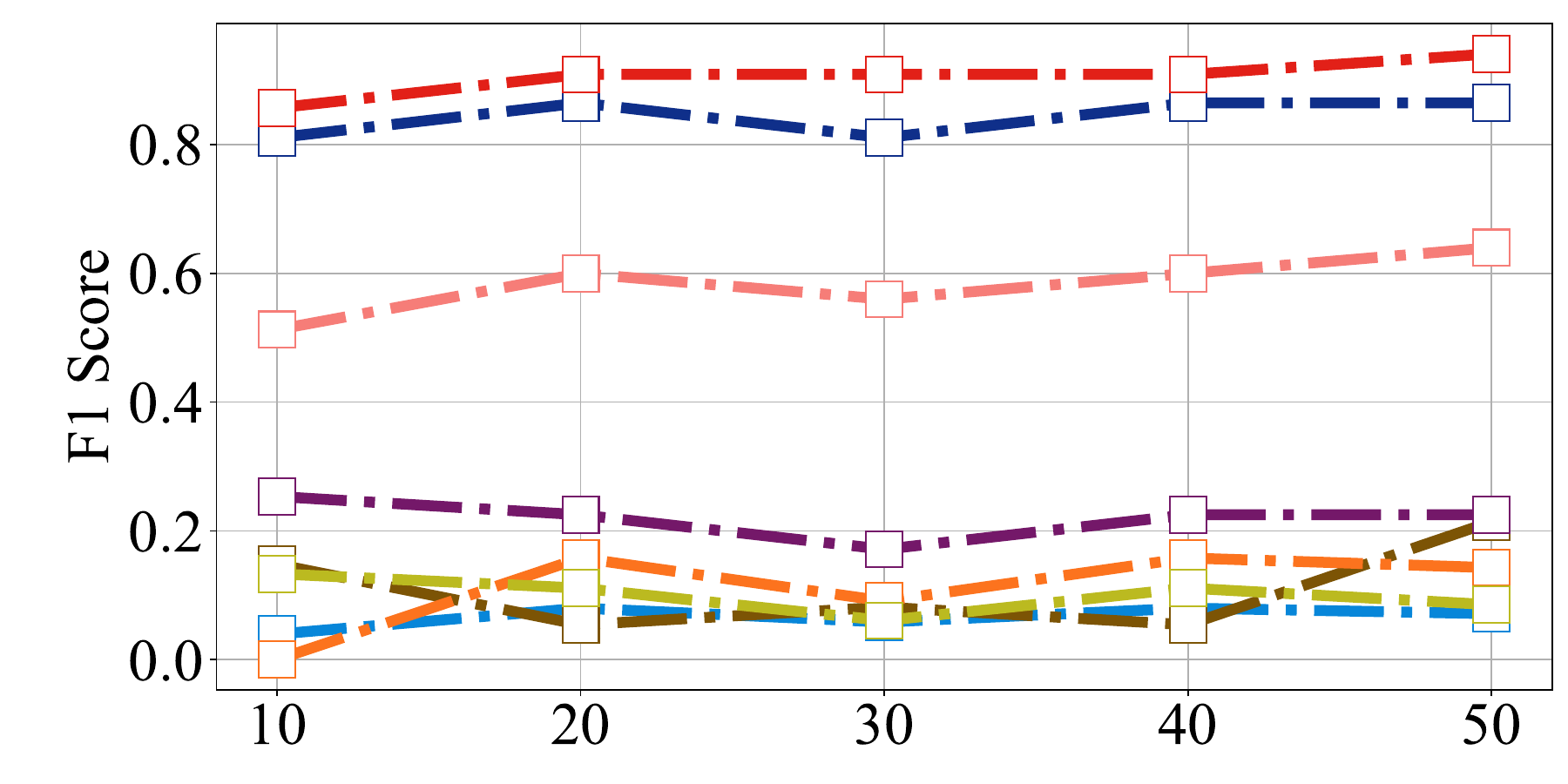}
	}
	\hfill
	\subfloat[Sensitivity to $\mu$ (F1 Score)]{
		\includegraphics[width=0.32\linewidth]{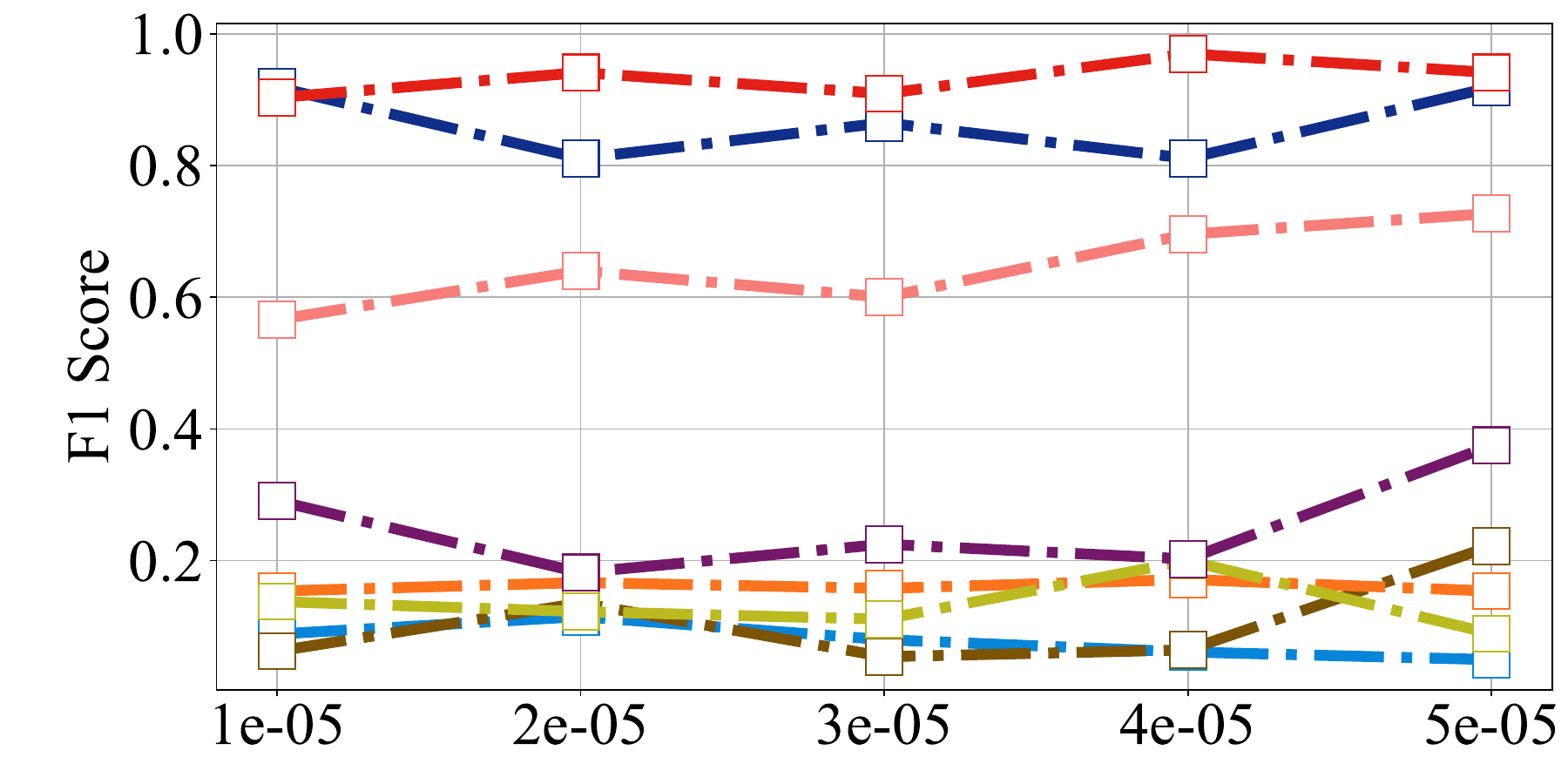}
	}

	\centering
	\scriptsize
	\subfloat[Sensitivity to $\alpha$ (F1 Score)]{
		\includegraphics[width=0.32\linewidth]{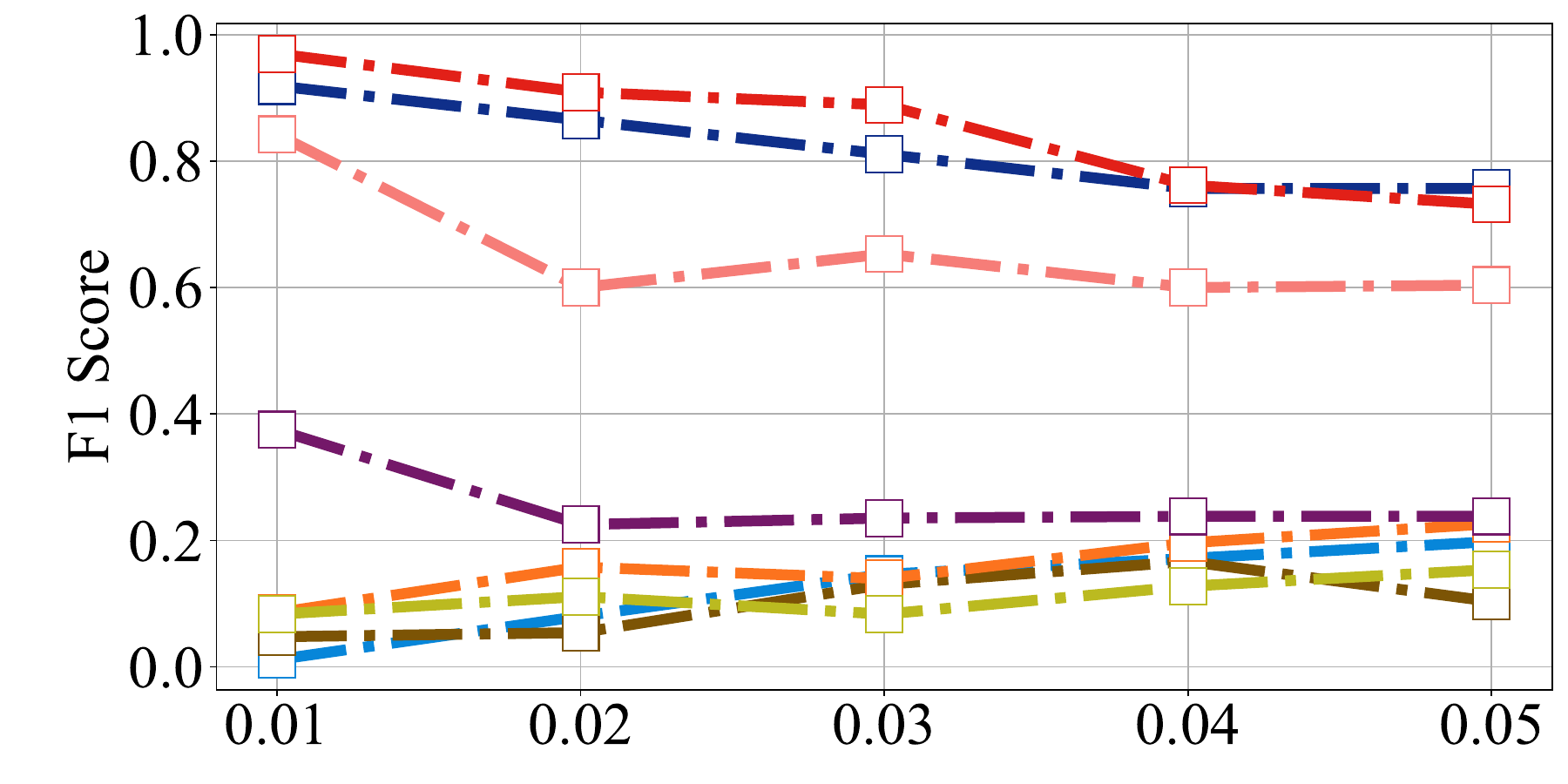}
	}
	\hfill
	\subfloat[Sensitivity to Period $T$ (F1 Score)]{
		\includegraphics[width=0.32\linewidth]{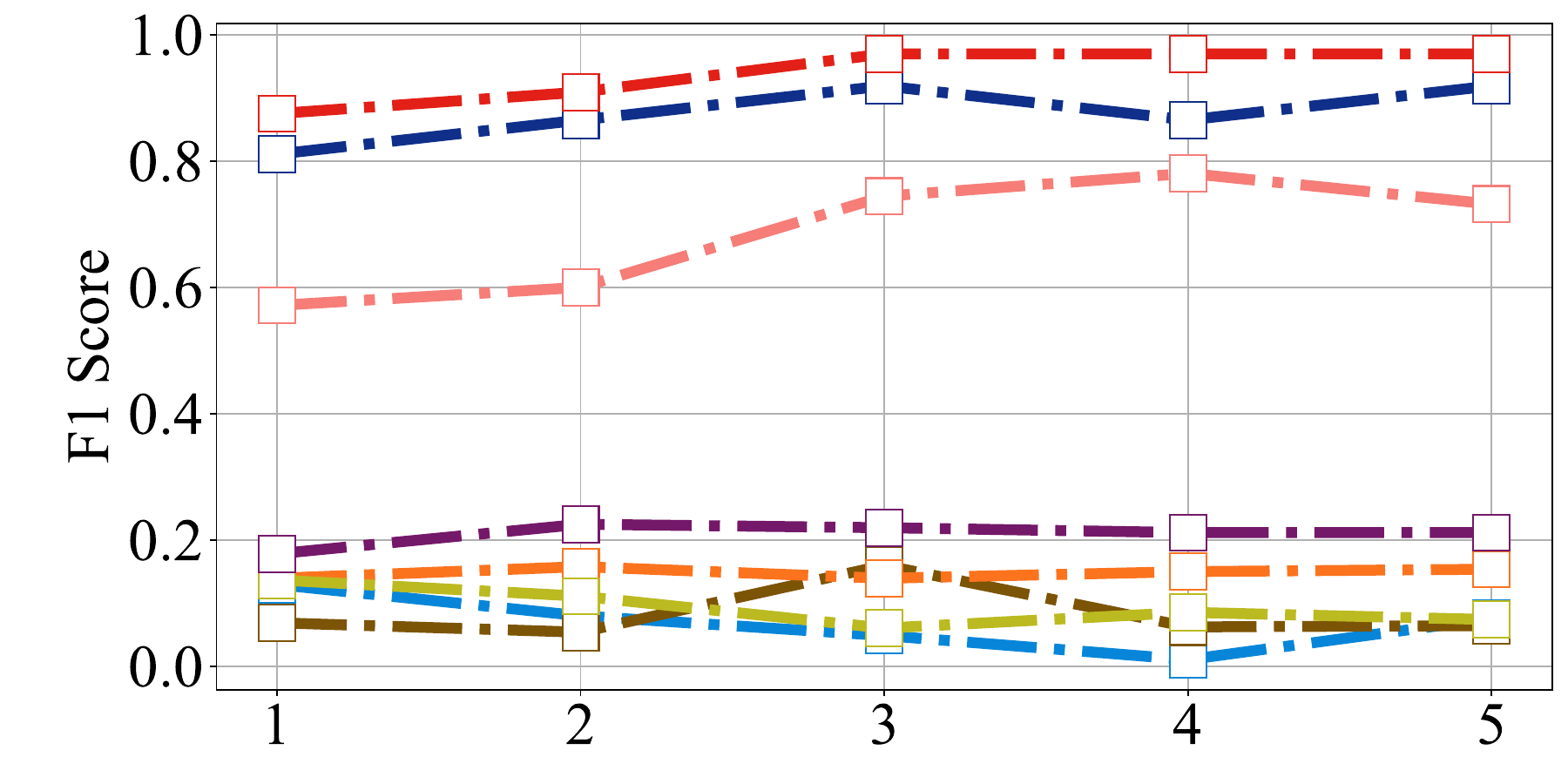}
	}
	\hfill
	\subfloat[Sensitivity to $\Delta$ (F1 Score)]{
		\includegraphics[width=0.32\linewidth]{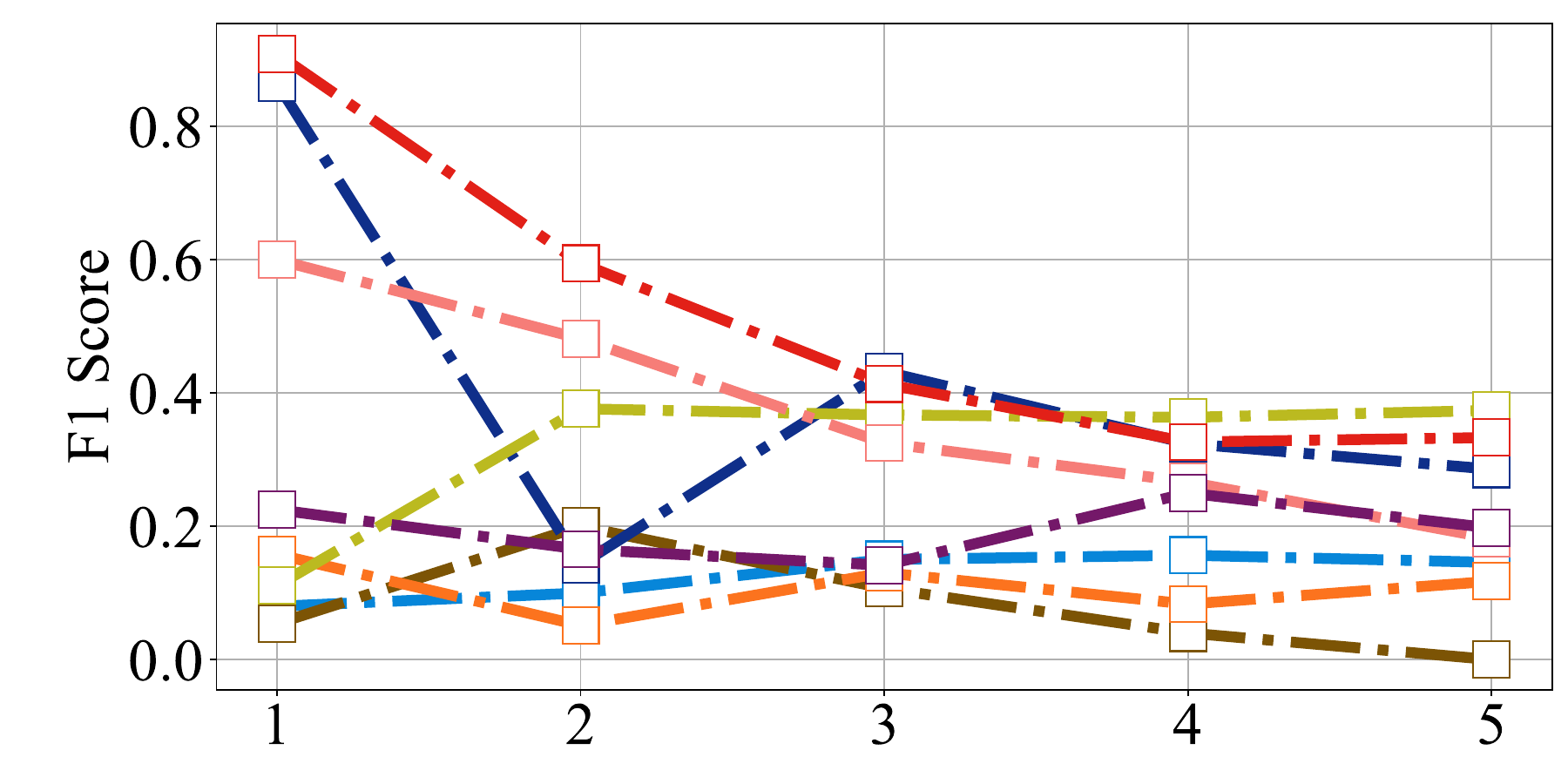}
	}

	\centering
	\scriptsize
	\subfloat[Sensitivity to Event Types (SHD)]{
		\includegraphics[width=0.32\linewidth]{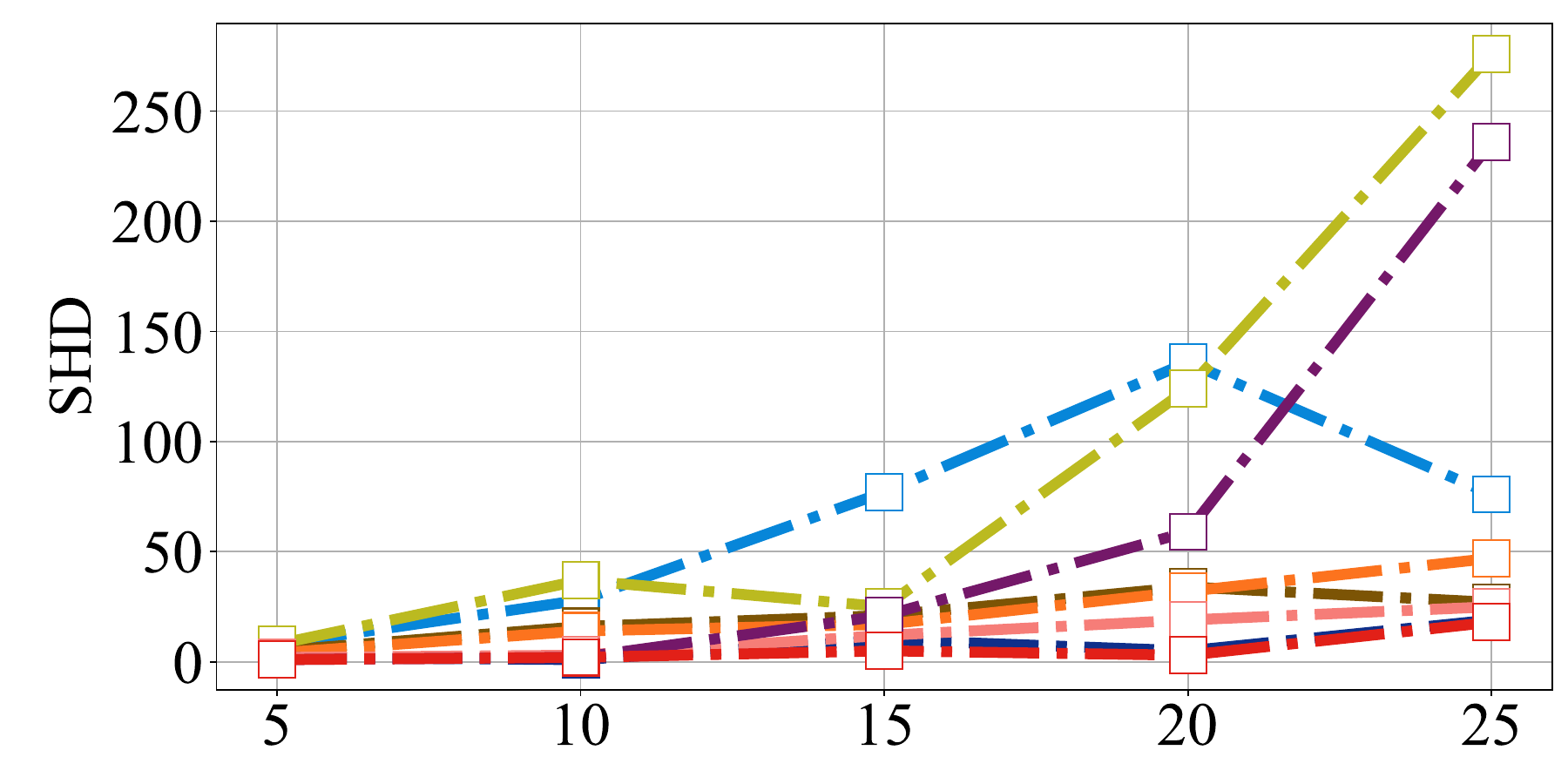}
	}
	\hfill
	\subfloat[Sensitivity to Devices (SHD)]{
		\includegraphics[width=0.32\linewidth]{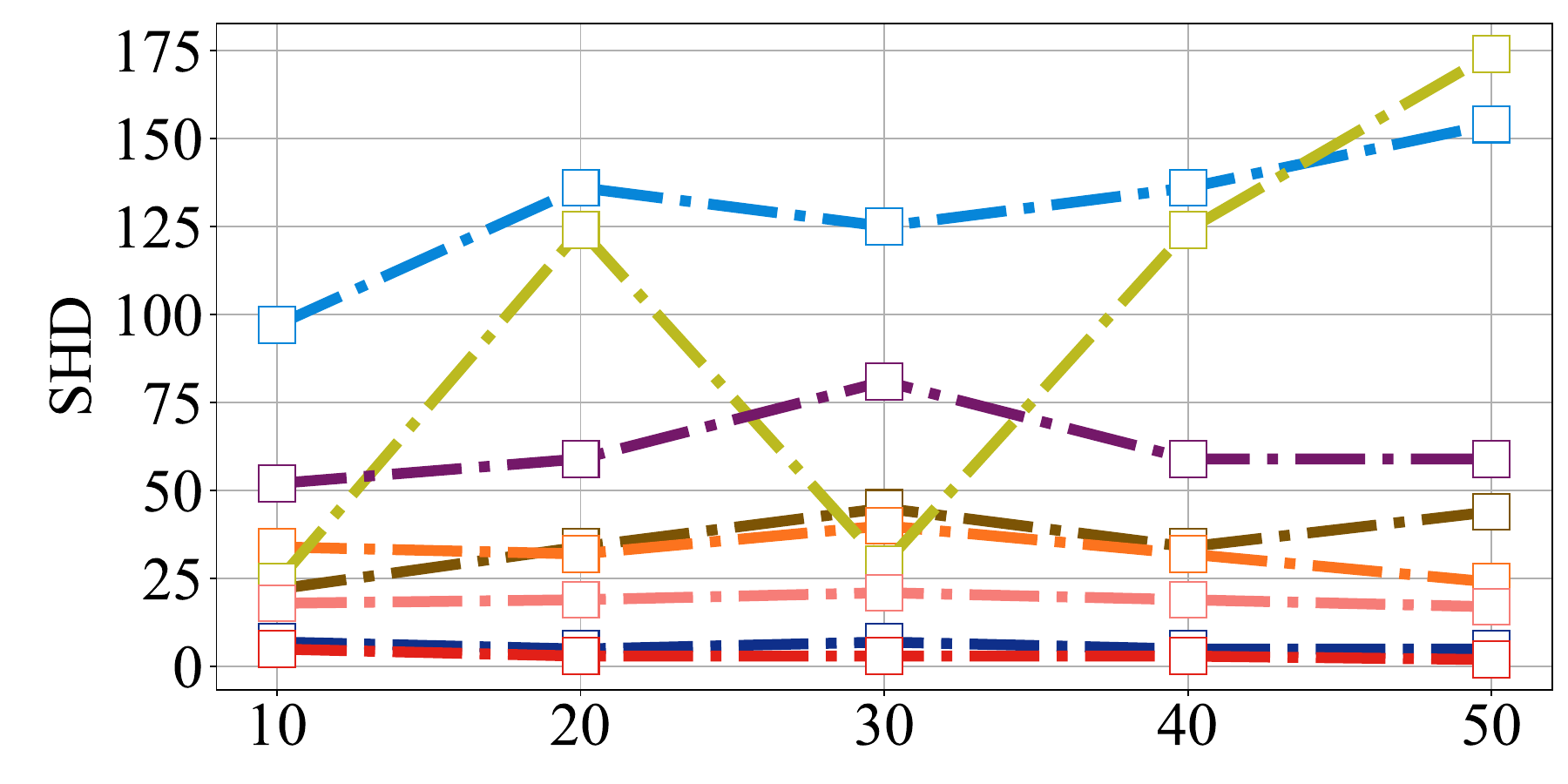}
	}
	\hfill
	\subfloat[Sensitivity to $\mu$ (SHD)]{
		\includegraphics[width=0.32\linewidth]{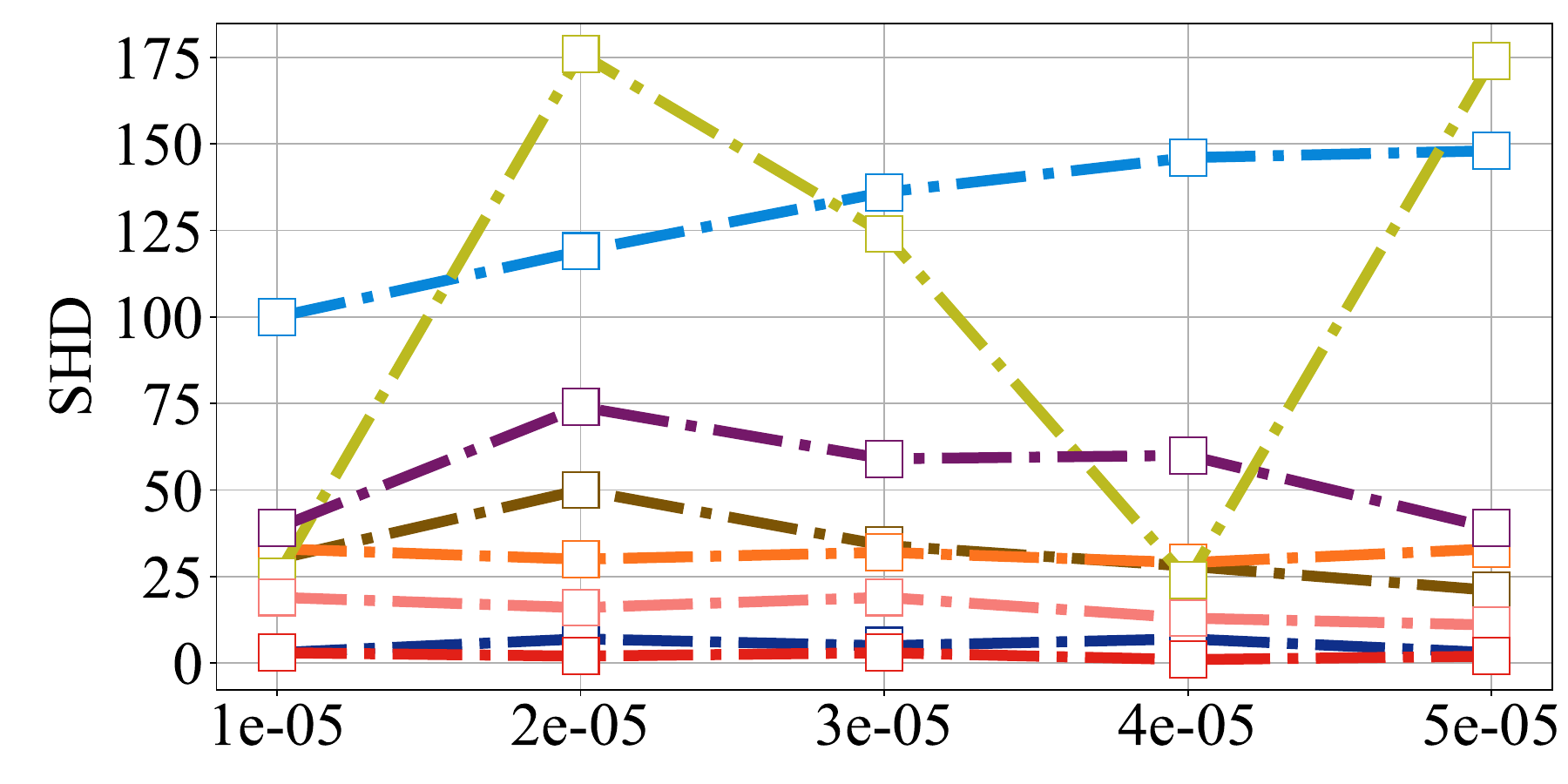}
	}

	\centering
	\scriptsize
	\subfloat[Sensitivity to $\alpha$ (SHD)]{
		\includegraphics[width=0.32\linewidth]{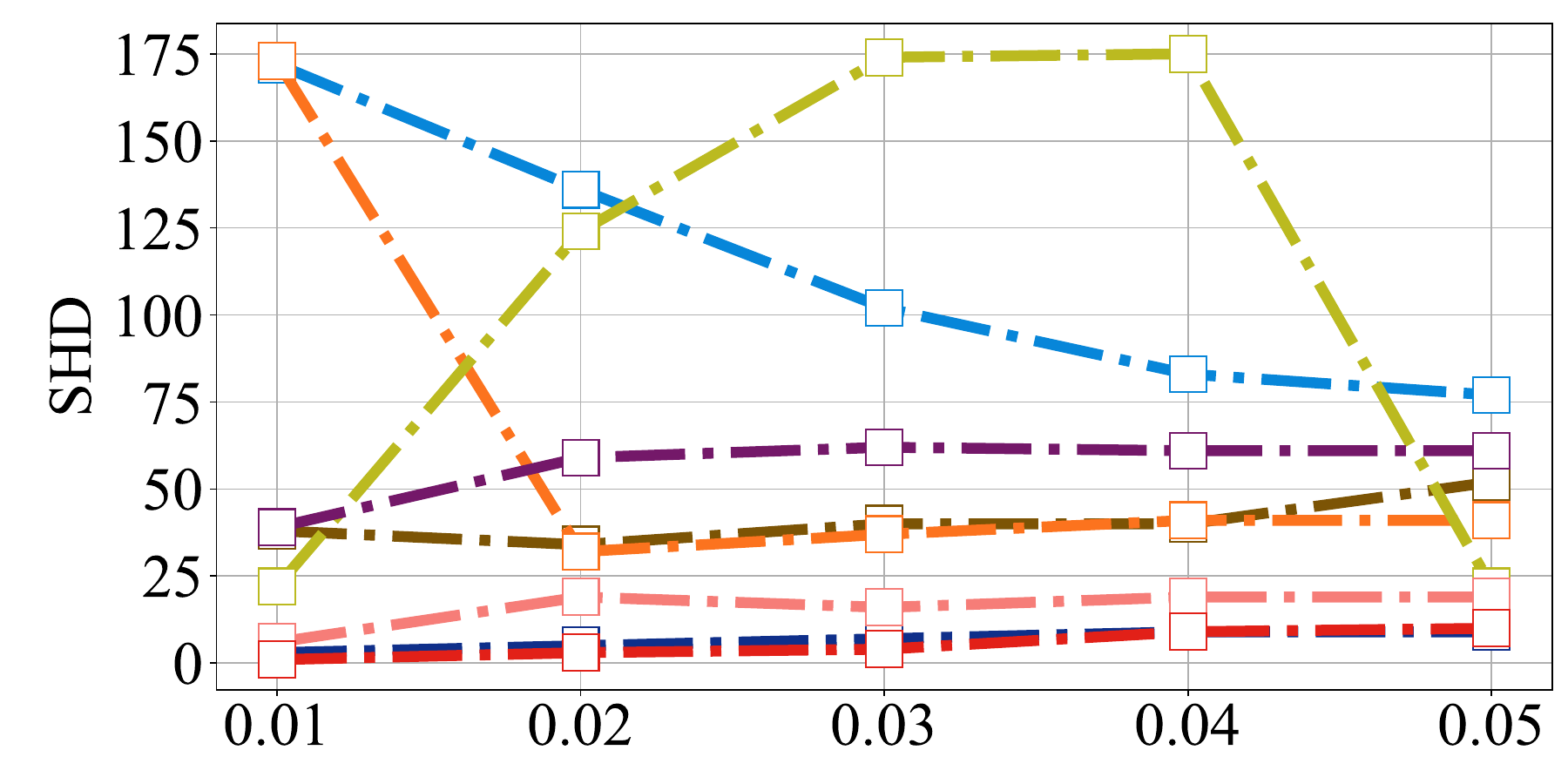}
	}
	\hfill
	\subfloat[Sensitivity to Period $T$ (SHD)]{
		\includegraphics[width=0.32\linewidth]{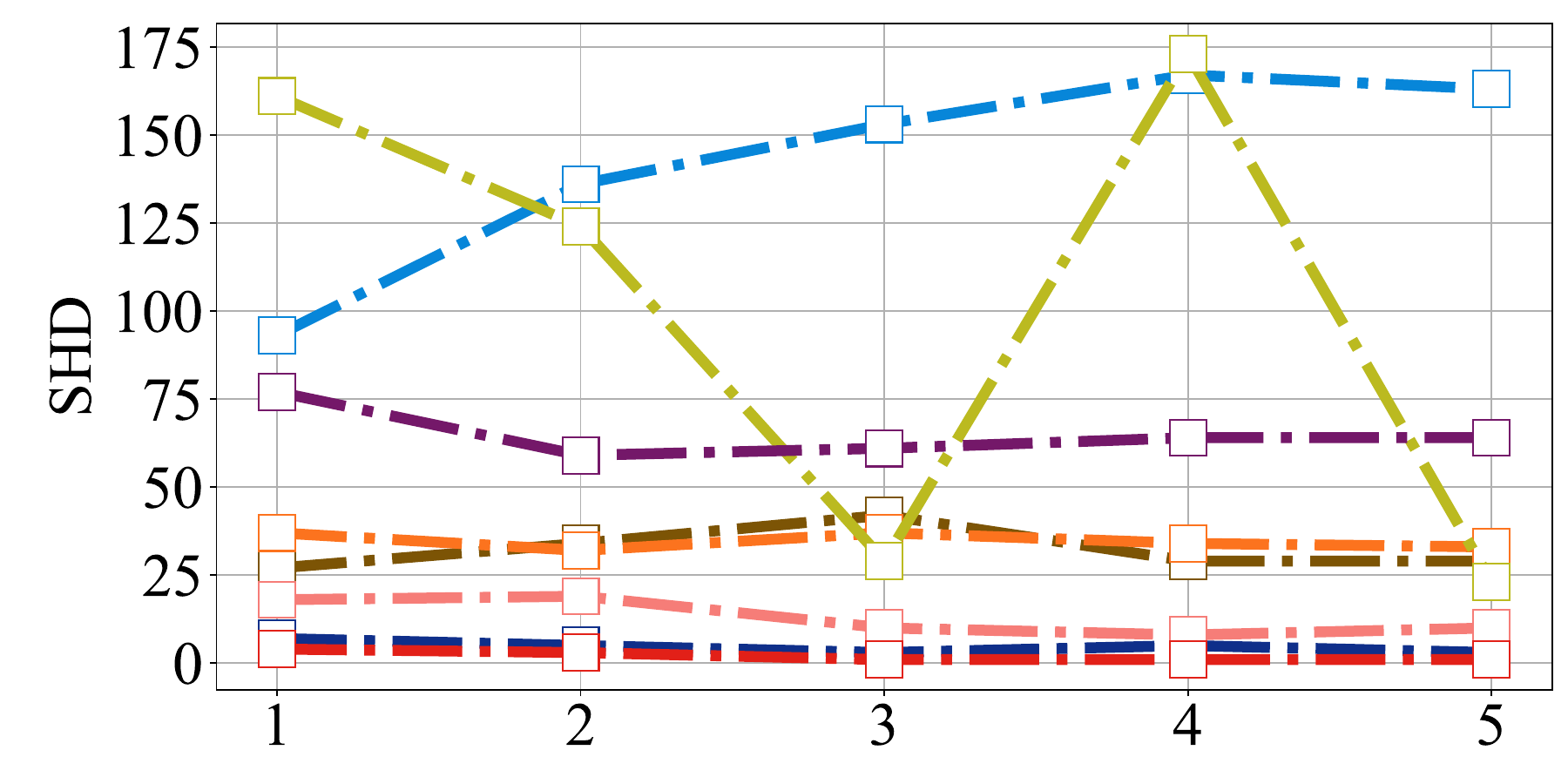}
	}
	\hfill
	\subfloat[Sensitivity to $\Delta$ (SHD)]{
		\includegraphics[width=0.32\linewidth]{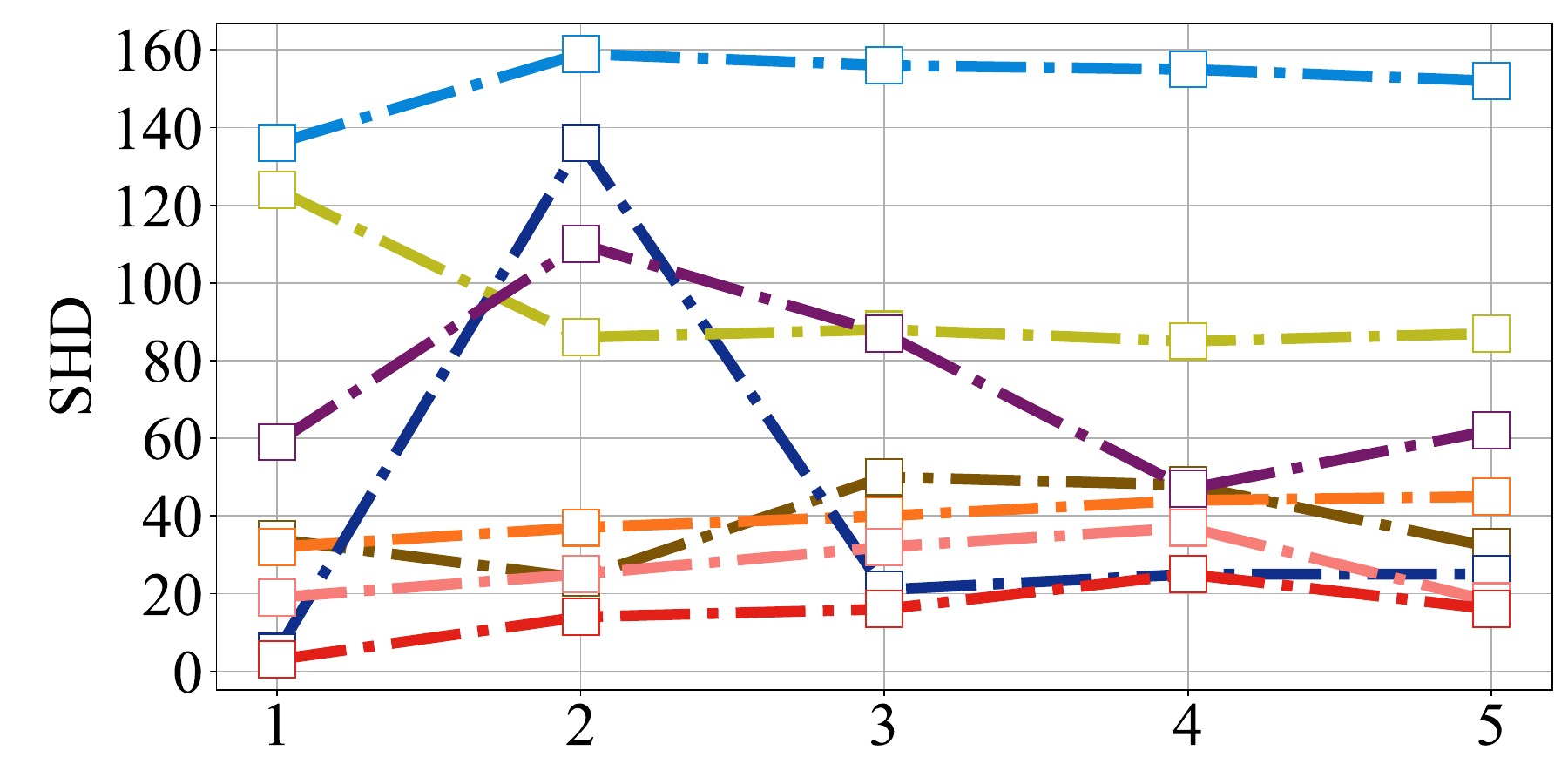}
	}

    \centering
    \scriptsize
	\subfloat{
	\includegraphics[width=0.55\linewidth]{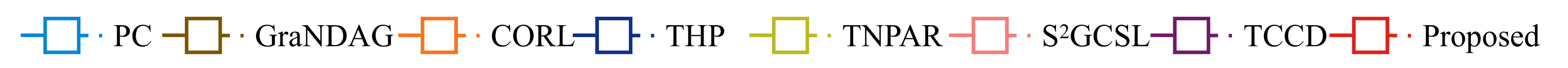}
    }
\caption{Performance comparison of different causal discovery methods on synthetic datasets.}
\label{env:f1shd}
\end{figure*}

To evaluate the causal discovery performance of our proposed method on synthetic telecommunication network event data, we conduct experiments using a controlled variable approach, systematically varying six parameters in the data generation process: the number of event types, the number of devices, the baseline intensity $\mu$, the excitation strength $\alpha$, the generation period $T$, and the discretization interval $\Delta$. We record the F1 Score and SHD of each method under various parameter configurations. As shown in Figure \ref{env:f1shd}, PMCDTrace demonstrates the best or near-best comprehensive performance in most scenarios, particularly achieving the leading F1 Score and the lowest SHD values, indicating its superiority in both accuracy and structural fidelity.

PMCDTrace maintains a leading position across the vast majority of parameter variations. As the number of devices increases, its F1 Score consistently remains above $0.85$, significantly outperforming THP and TCCD. When varying the baseline intensity $\mu$ and excitation strength $\alpha$, PMCDTrace consistently maintains a high F1 Score around $0.90$, reaching $0.97$ at $\mu = 4\times 10^{-5}$, demonstrating strong robustness to intensity parameters. As the generation period $T$ increases, the F1 Score of PMCDTrace steadily rises from $0.88$ to $0.97$, whereas other methods such as PC and GraNDAG generally decline or exhibit significant fluctuations. Notably, as the discretization interval $\delta$ increases, the F1 Score of all methods show a decreasing trend, with PMCDTrace dropping from $0.91$ to $0.33$, indicating that overly coarse time granularity poses challenges for causal discovery.

PMCDTrace achieves the lowest SHD values under most experimental conditions, demonstrating that its predicted causal graphs exhibit the smallest structural discrepancies from the ground truth. As the number of event types and devices increases, the SHD of PMCDTrace remains below $5$, significantly outperforming advanced methods such as S\textsuperscript{2}GCSL and TCCD. When varying $\mu$, $\alpha$, and $T$, the SHD of PMCDTrace typically ranges between $1$ and $4$, achieving an optimal value of $1$ at $\alpha = 0.01$. Although THP also performs well, its SHD sharply rises to $136$ as $\delta$ increases, exhibiting inferior robustness compared to PMCDTrace. As $\delta$ increases, the SHD of PMCDTrace rises from $3$ to $25$; while this remains superior to most baselines, it still suggests that coarser temporal granularity affects structural recovery to some extent. In summary, PMCDTrace exhibits a clear advantage in maintaining low SHD, consistently producing near-ground-truth graph structures even as the complexity of the true structure increases.

In conclusion, PMCDTrace comprehensively outperforms existing mainstream causal discovery methods on synthetic datasets. It not only achieves high and stable F1 Score but also significantly lower SHD values compared to baseline methods, demonstrating excellent accuracy, structural recovery capability, and parameter robustness. The method maintains strong performance across varying network scales, numbers of event types, and temporal granularities, confirming its effectiveness and generalization potential for causal inference in real-world telecommunication network scenarios.

\subsection{RQ2: Performance Comparison on Real-World Datasets}

\begin{table*}[ht]
\newcommand{\tabincell}[2]{\begin{tabular}{@{}#1@{}}#2\end{tabular}}
\centering
\begin{threeparttable}
\setlength\tabcolsep{6.0pt}
\caption{Performance comparison of different causal discovery methods on real datasets}
\begin{tabular}{cccccccccccccc}
\toprule  
\multirow{2}*{Types} &\multirow{2}*{Methods}& \multicolumn{4}{c}{18V-55N-Wireless} & \multicolumn{4}{c}{24V-439N-Microwave}& \multicolumn{4}{c}{25V-474N-Microwave}\\
\cline{3-14}~ &~ &PRE & REC & F1 & SHD& PRE & REC & F1 & SHD& PRE & REC & F1 & SHD\\
\midrule
\multirow{3}*{\makecell[c]{Without Topology}}&PC	&0.3307&	0.6087&	0.4286&	88	&0.3243	&0.2628	&0.2903	&142	&0.2821&	0.2230	&0.2491	&162\\
~&GraNDAG&	0.3333&	0.1014&	0.1556&	71&	0.2500&	0.0438&	0.0745&	144&	0.4667&	0.0473	&0.0859&	143\\
~&CORL&	0.1250&	0.0580&	0.0792&	83	&0.1176&	0.0146&	0.0260&	140&	0.2000&	0.0338&	0.0578&	158\\
\bottomrule
\multirow{5}*{\makecell[c]{With Topology}}&THP&	0.6667&	0.1449&	0.2381&	62&	0.6000	&0.0657&	0.1184	&129&	0.6000&	0.0811&	0.1429&	136\\
~&TNPAR&	0.2598&	0.7681&	0.3883&	100	&0.2524	&0.9781	&0.4012	&142	&0.2462	&0.9932	&0.3946	&153\\
~&TCCD	&0.2407&	0.9420&	0.3835&	88&	0.2394&	0.9489&	0.3824&	146&	0.2467&	1.0000&	0.3957&	152\\
~&S$^2$GCSL&	0.4674&	0.6232&	0.5342&	55&	0.2407&	0.9489&	0.3840&	146&	0.2467&	1.0000&	0.3957&	152\\
~&\textbf{Proposed}	&\textbf{0.5250}&	\textbf{0.6087}&	\textbf{0.5638}	&\textbf{39}	&\textbf{0.2604}&	\textbf{0.9124}&	\textbf{0.4052}&	\textbf{132}	&\textbf{0.2576}&	\textbf{0.9122}&	\textbf{0.4018}&	\textbf{145}\\
\bottomrule
\end{tabular}
\label{table1}
\end{threeparttable}
\end{table*}

To validate the generalization ability and practical utility of our proposed method on real-world metropolitan cellular network alarm data, we conducted comprehensive comparisons with seven baseline methods on the 18V-55N-Wireless, 24V-439N-Microwave, and 25V-474N-Microwave datasets. As presented in Table \ref{table1}, PMCDTrace achieves the highest F1 Score across all three datasets and obtains the lowest SHD in most cases, demonstrating its superior comprehensive performance in complex real-world scenarios.

PMCDTrace achieves the best performance on all three real-world datasets, exhibiting excellent generalization capability. On the 18V-55N-Wireless dataset, it attains an F1 Score of $0.5638$, significantly outperforming S\textsuperscript{2}GCSL. On the two larger microwave datasets, PMCDTrace maintains its leading F1 Score and substantially surpasses methods such as THP that focus solely on topological information. Notably, although TNPAR and TCCD achieve Recall close to $1.0$, their low Precision results in relatively low F1 Score. These experimental results indicate that PMCDTrace effectively adapts to real-world data of varying scales and sparsity levels.

PMCDTrace also demonstrates a clear advantage in the accuracy of structural recovery. On the 18V-55N-Wireless dataset, its SHD is only $39$, considerably lower than THP and S\textsuperscript{2}GCSL, indicating that its predicted graph exhibits the smallest structural discrepancy from the ground truth. On larger-scale datasets, although the absolute SHD values of PMCDTrace increase, they remain lower than or comparable to most baselines. Particularly when the network scales up to hundreds of devices, PMCDTrace exhibits relatively robust structural recovery. It is worth noting that methods such as TNPAR and TCCD, despite achieving extremely high Recall, suffer from elevated SHD due to covering true edges through numerous false positives, resulting in structural redundancy. In contrast, PMCDTrace performs better in controlling the false positive rate, thereby achieving lower SHD.

In summary, PMCDTrace comprehensively outperforms existing baselines on real-world telecommunication network alarm datasets. It not only maintains a leading F1 Score but also demonstrates superior structural fidelity in terms of SHD, confirming its ability to effectively adapt to complex real-world scenarios with strong generalization capacity and practical potential. Although there remains room for improvement in SHD on ultra-large-scale networks, the overall performance of PMCDTrace sufficiently validates its value as a reliable causal discovery tool, providing robust support for root cause analysis in network operations.

\subsection{RQ3: Efficiency Comparison of Causal Discovery Methods}
\begin{figure*}[!htb]
	\centering
	\scriptsize
	\subfloat[Time for Causal Discovery on Synthetic Datasets]{
		\includegraphics[width=0.48\linewidth]{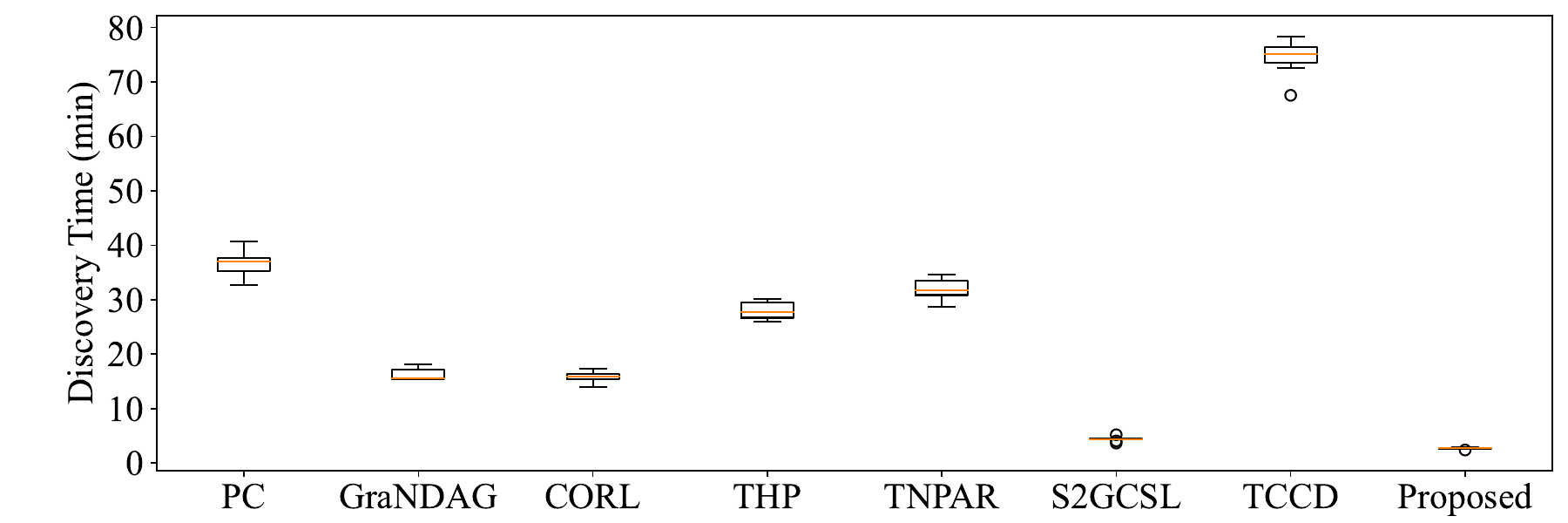}
	}
	\hfill
	\subfloat[Time for Causal Discovery on 18V-55N-Wireless]{
		\includegraphics[width=0.48\linewidth]{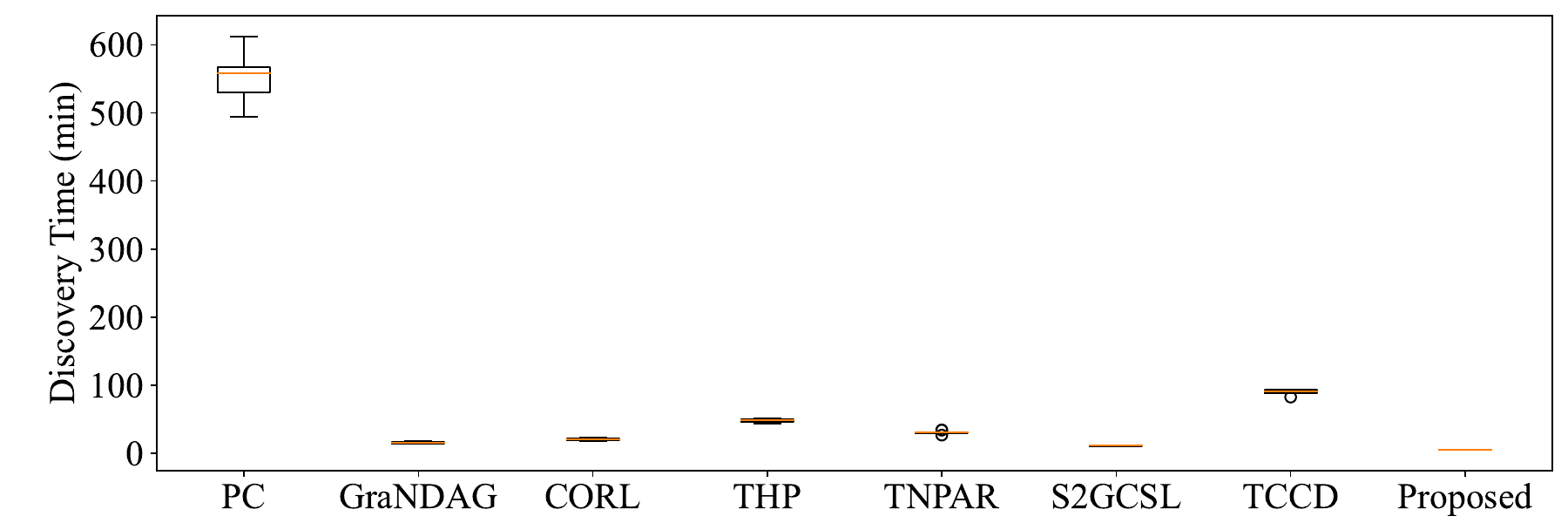}
	}
    
    \centering
	\scriptsize
	\hfill
	\subfloat[Time for Causal Discovery on 24V-439N-Microwave]{
		\includegraphics[width=0.48\linewidth]{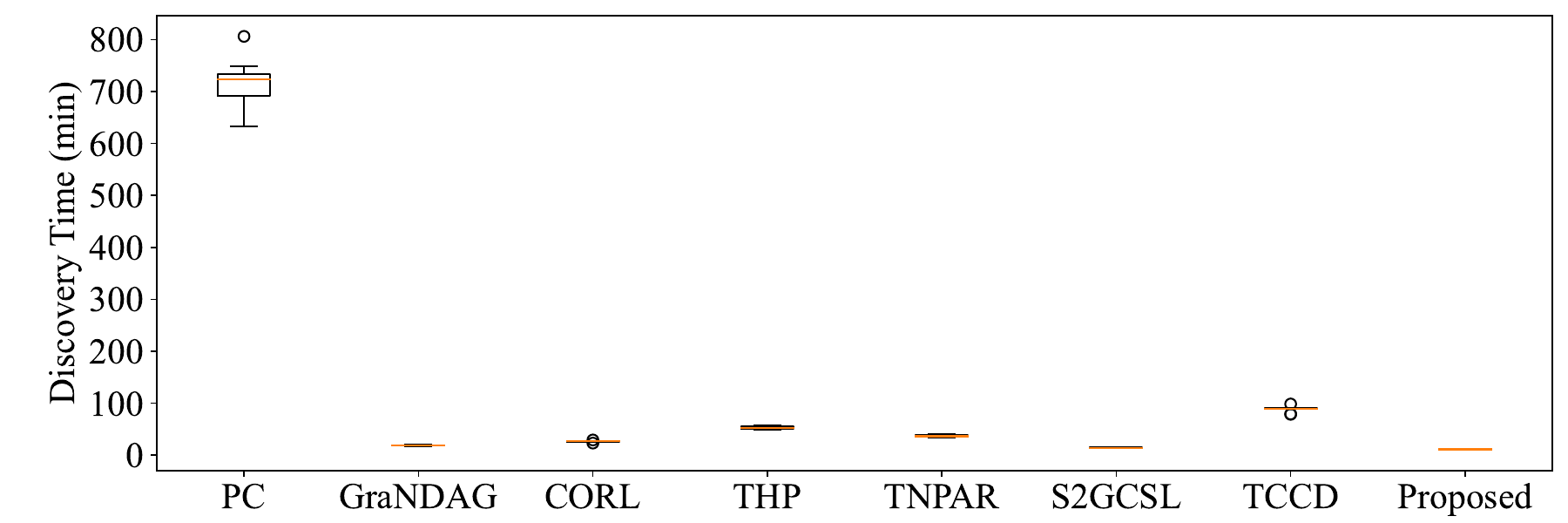}
	}
    \hfill
	\subfloat[Time for Causal Discovery on 25V-474N-Microwave]{
		\includegraphics[width=0.48\linewidth]{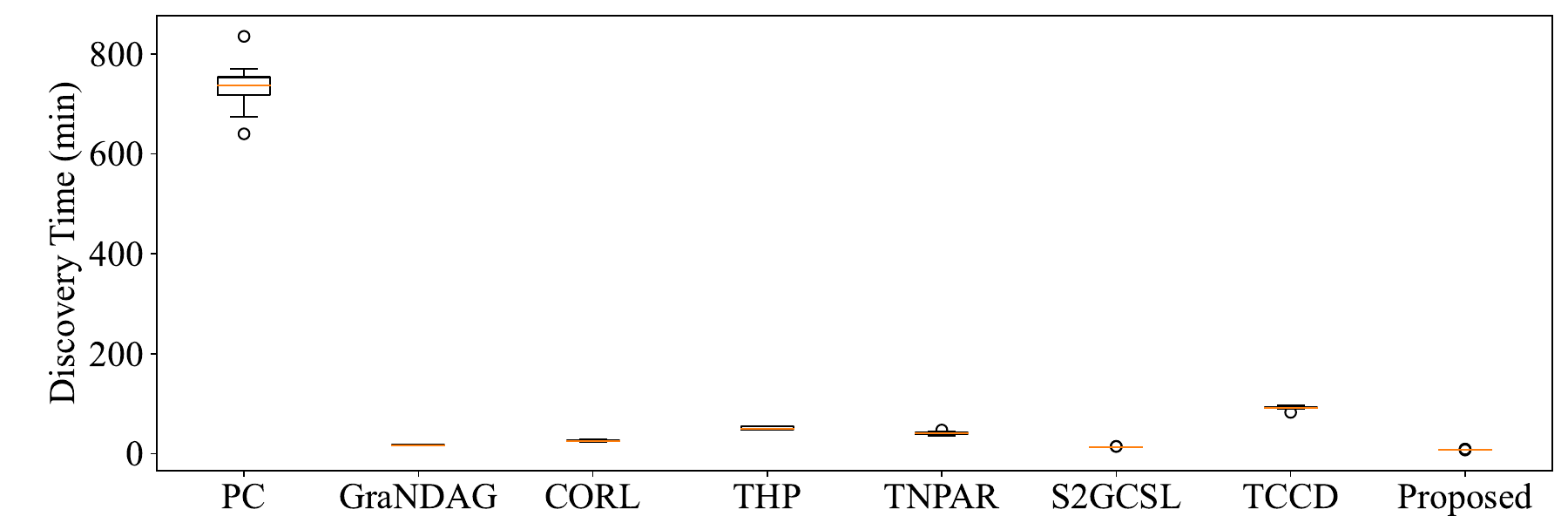}
	}
\caption{Comparison of Causal Discovery Efficiency for Different Methods}
\label{env:f2}
\end{figure*}

To evaluate the causal discovery efficiency of the proposed PMCDTrace method on both synthetic and real-world telecommunication network datasets, we conducted ten independent runs on one synthetic dataset with default parameters and on the three real-world datasets. We recorded the wall-clock time required by each method to complete causal discovery, presenting the results as box plots. As illustrated in Figure \ref{env:f2}, PMCDTrace exhibits the best time efficiency across all datasets, with running times significantly lower than those of other baseline methods.

On the synthetic dataset, PMCDTrace requires only 3 minutes, which is faster than the next best method, S\textsuperscript{2}GCSL, while PC and TCCD take 38 minutes and 75 minutes, respectively. On the real-world dataset 18V-55N-Wireless, PMCDTrace completes in just 5 minutes, achieving a speedup of 112$\times$ over PC and 19$\times$ over TCCD. On 24V-439N-Microwave, PMCDTrace takes only 10 minutes, whereas PC requires up to 790 minutes. On 25V-474N-Microwave, PMCDTrace maintains a runtime of 5 minutes, compared to 750 minutes for PC and 130 minutes for TCCD. Notably, the runtime of PMCDTrace increases only modestly, from 3 minutes on the synthetic dataset to 10 minutes on the largest real-world dataset, while methods such as PC exhibit exponential growth in runtime. This demonstrates the excellent scalability and computational efficiency of PMCDTrace. Furthermore, when compared to the similarly efficient S\textsuperscript{2}GCSL, PMCDTrace retains a clear efficiency advantage despite the additional computational load introduced by modeling multi-scale interactions.

Overall, PMCDTrace demonstrates superior time efficiency in causal discovery, substantially outperforming both traditional methods and state-of-the-art efficiency-focused approaches. These results confirm that PMCDTrace can effectively handle large-scale telecommunication network data, meeting the demand for rapid response in real-world network operations.

\subsection{RQ4: Hyperparametric Sensitivity Analysis}

\begin{figure*}[!htb]
	\centering
	\scriptsize
	\subfloat[$\lambda$ on Synthetic Datasets]{
		\includegraphics[width=0.235\linewidth]{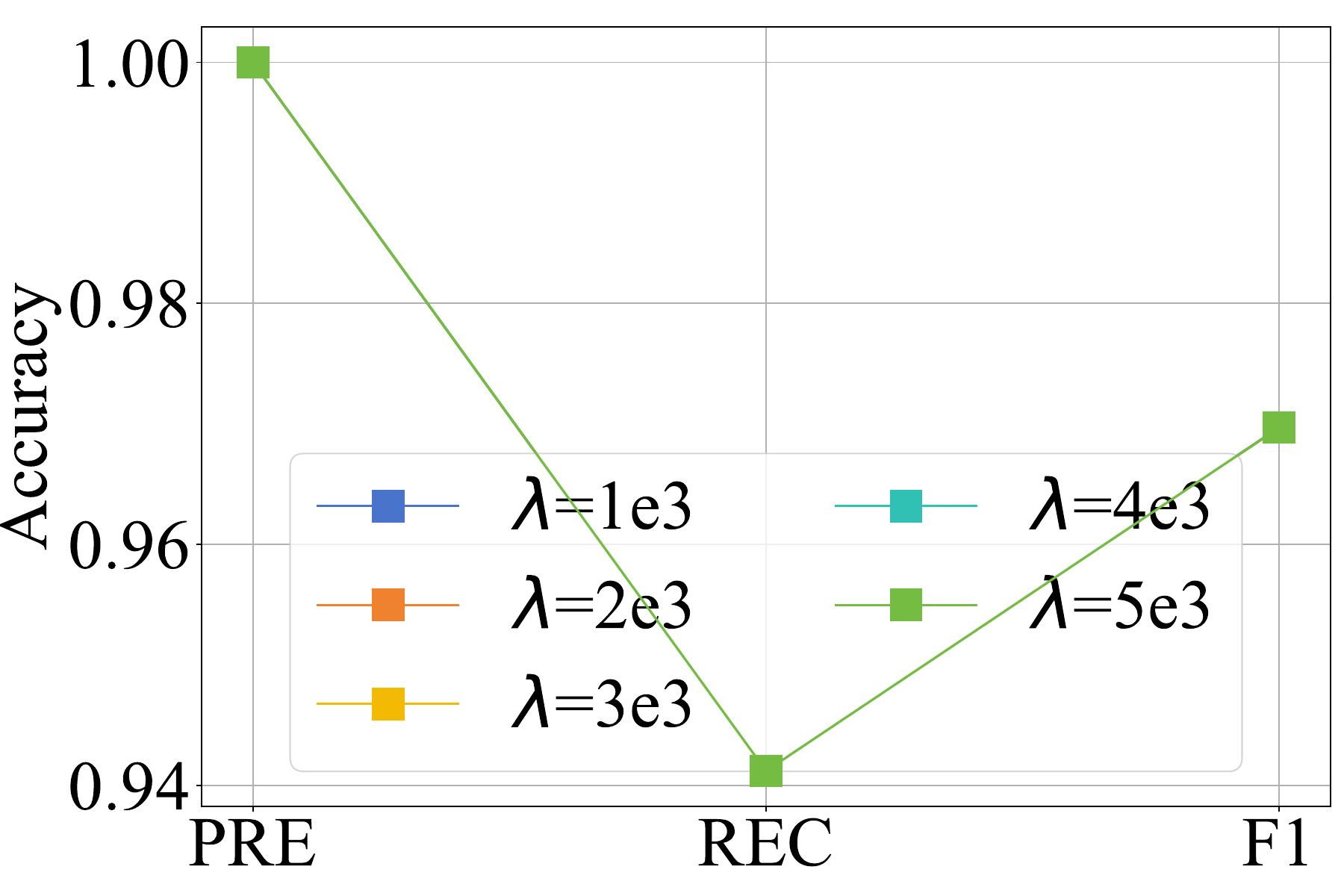}
	}
	\hfill
	\subfloat[$\lambda$ on 18V-55N-Wireless]{
		\includegraphics[width=0.235\linewidth]{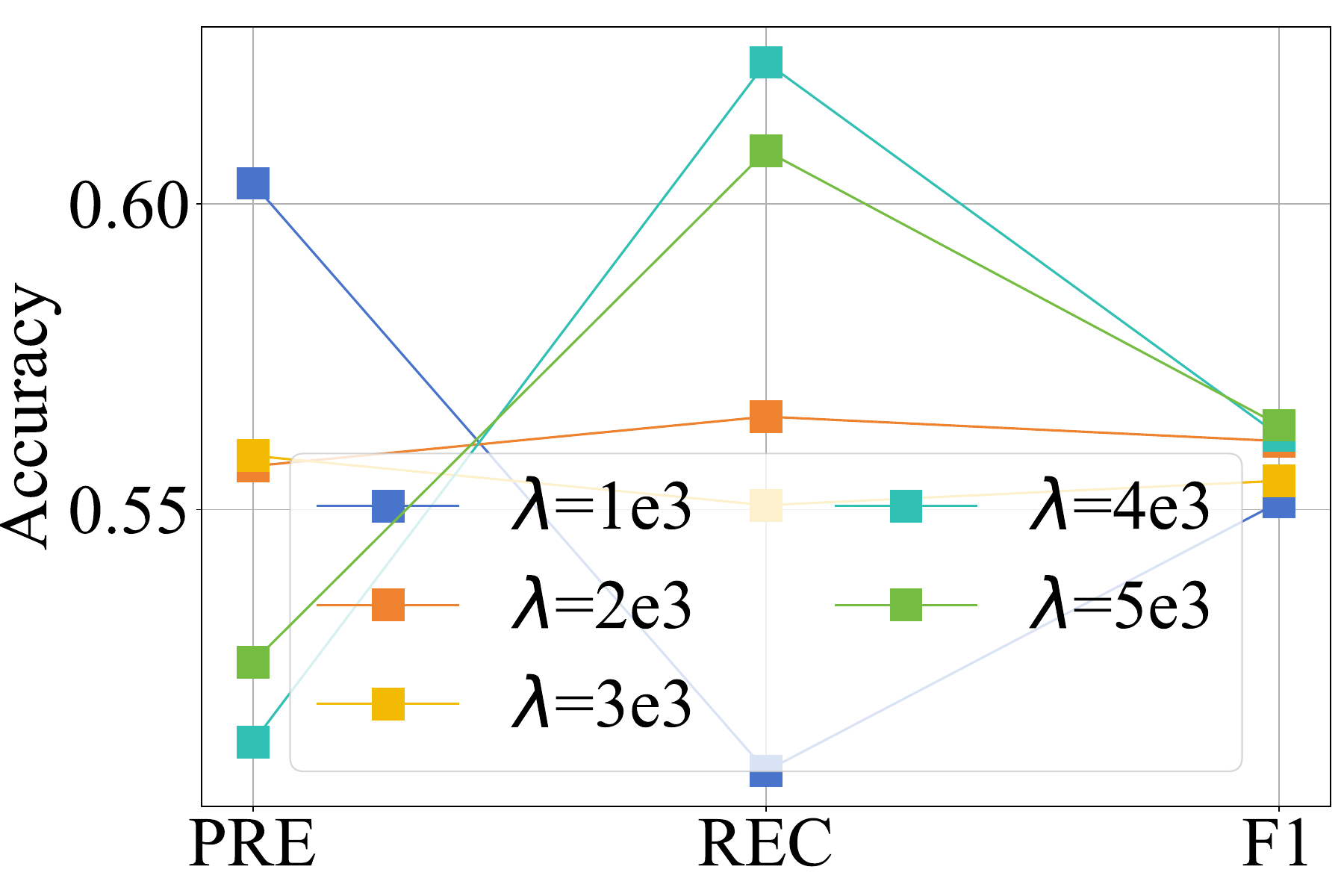}
	}
	\hfill
	\subfloat[$\lambda$ on 24V-439N-Microwave]{
		\includegraphics[width=0.235\linewidth]{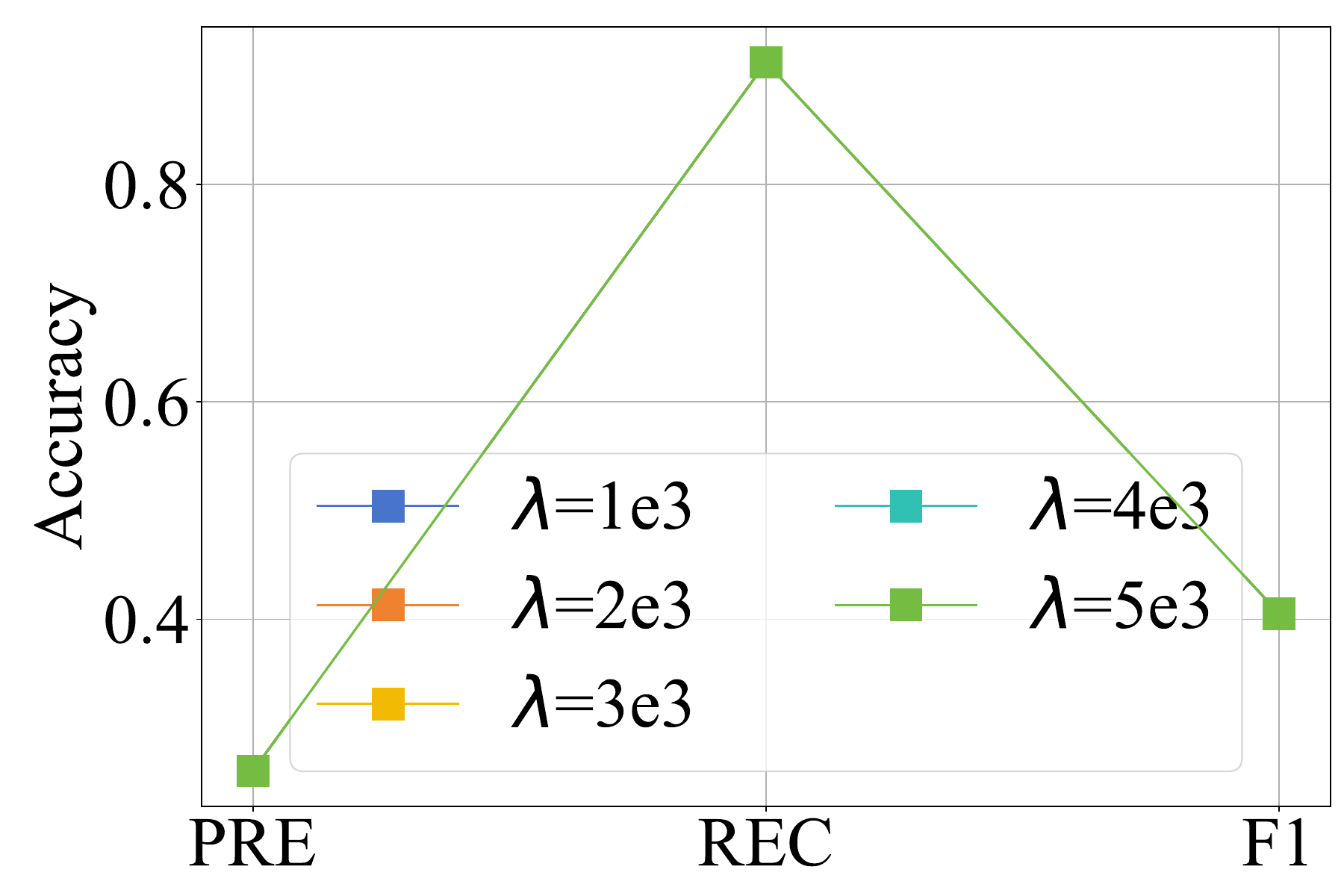}
	}
    \hfill
	\subfloat[$\lambda$ on 25V-474N-Microwave]{
		\includegraphics[width=0.235\linewidth]{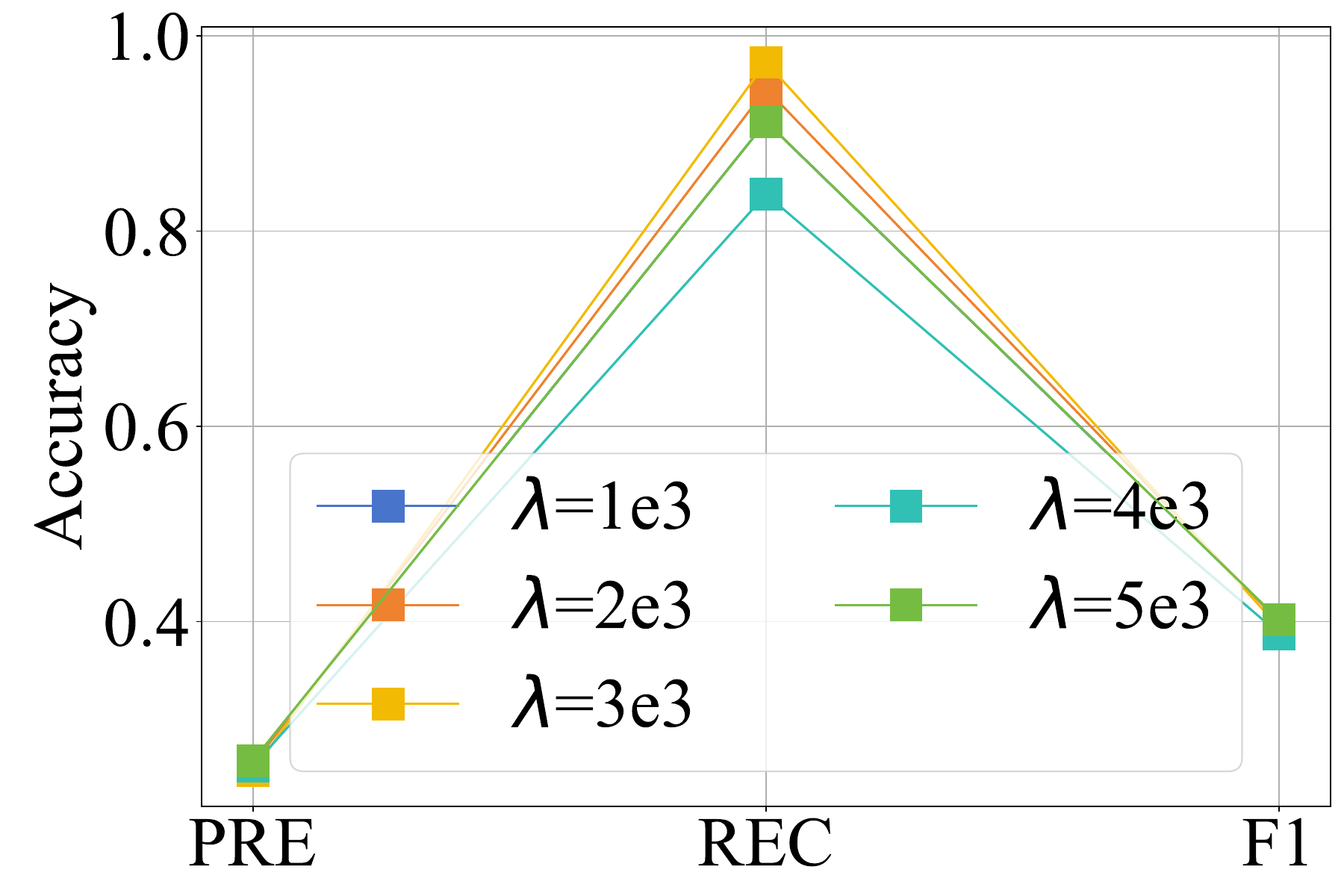}
	}

	\centering
	\scriptsize
	\subfloat[$\tau$ on Synthetic Datasets]{
		\includegraphics[width=0.235\linewidth]{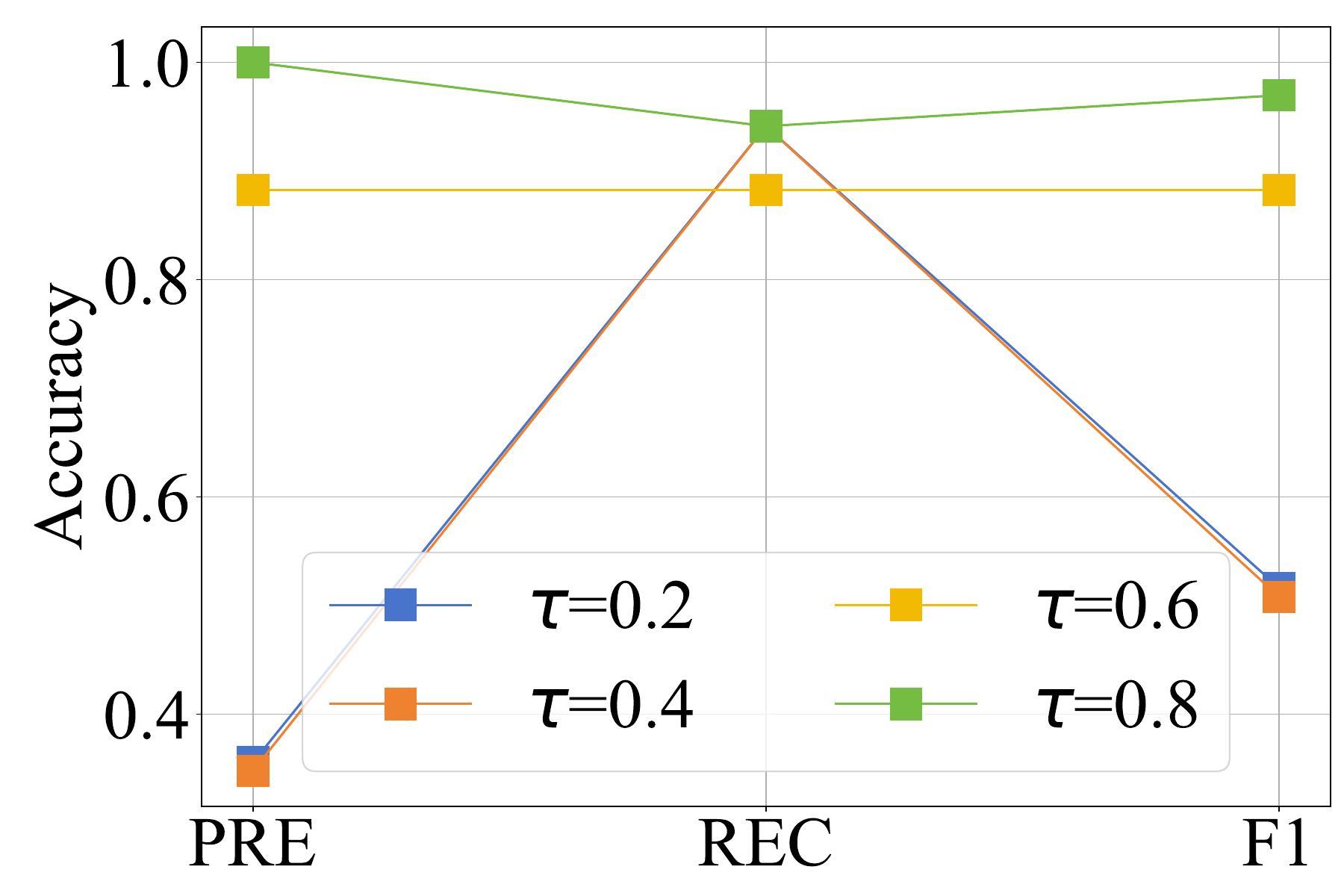}
	}
	\hfill
	\subfloat[$\tau$ on 18V-55N-Wireless]{
		\includegraphics[width=0.235\linewidth]{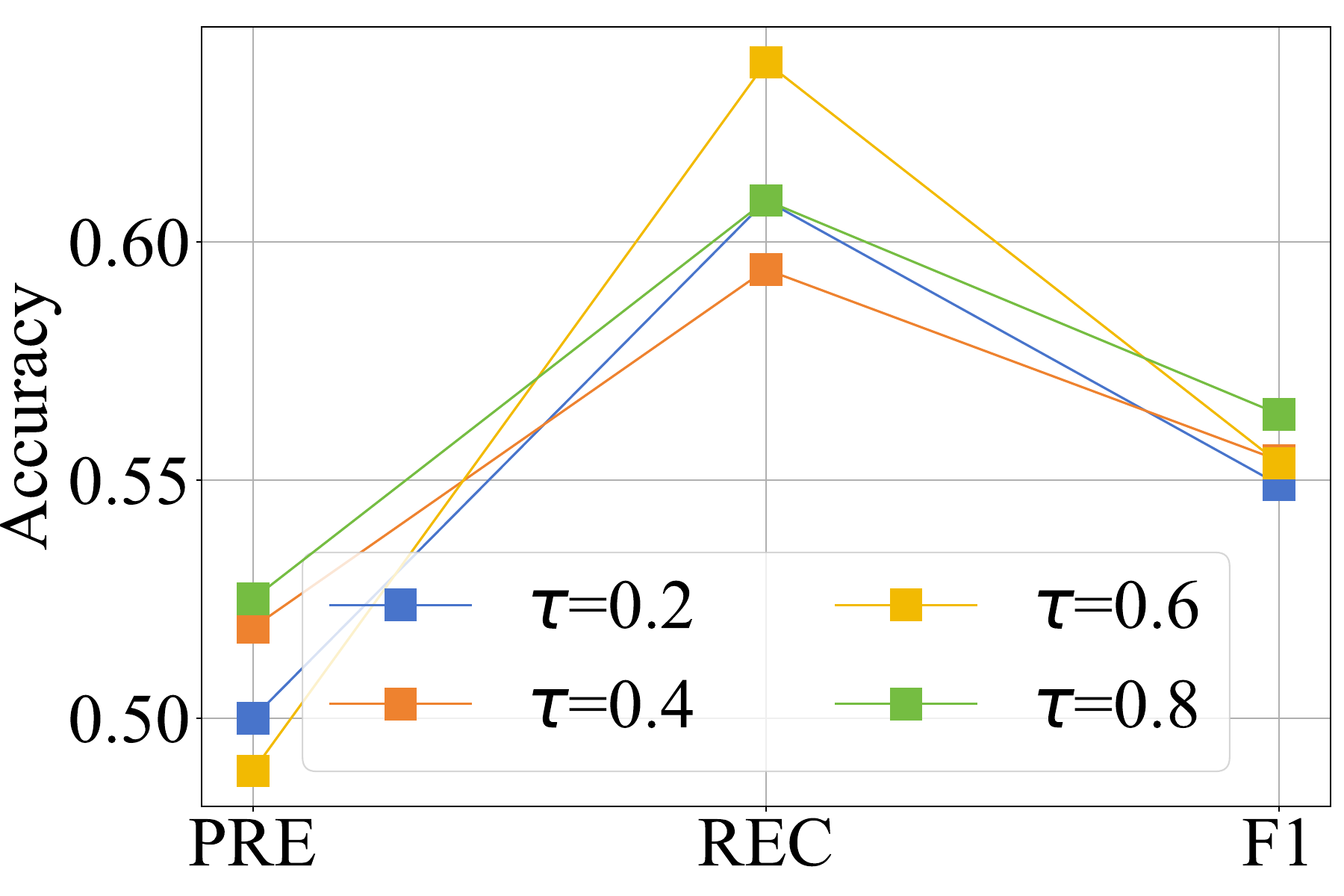}
	}
	\hfill
	\subfloat[$\tau$ on 24V-439N-Microwave]{
		\includegraphics[width=0.235\linewidth]{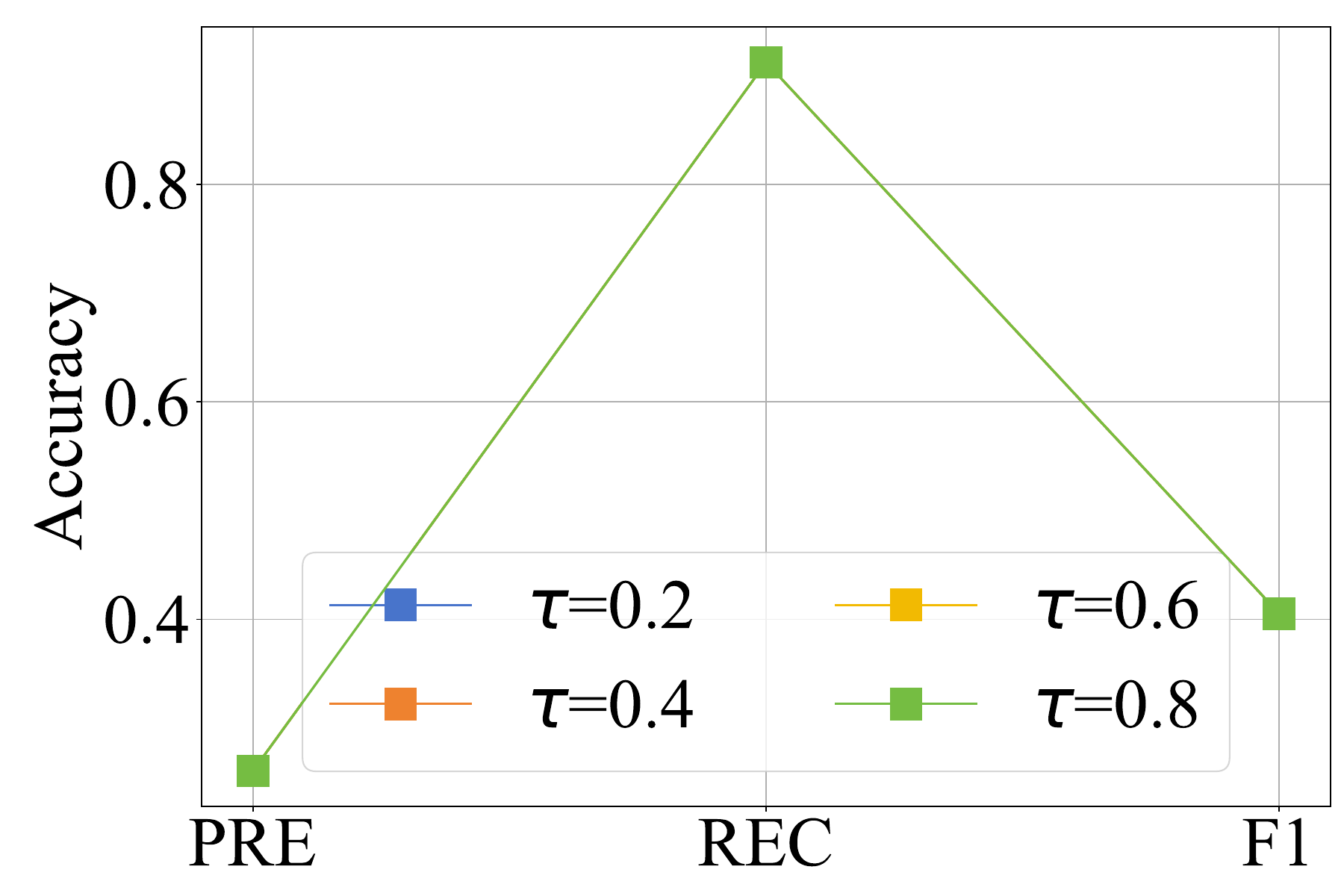}
	}
    \hfill
	\subfloat[$\tau$ on 25V-474N-Microwave]{
		\includegraphics[width=0.235\linewidth]{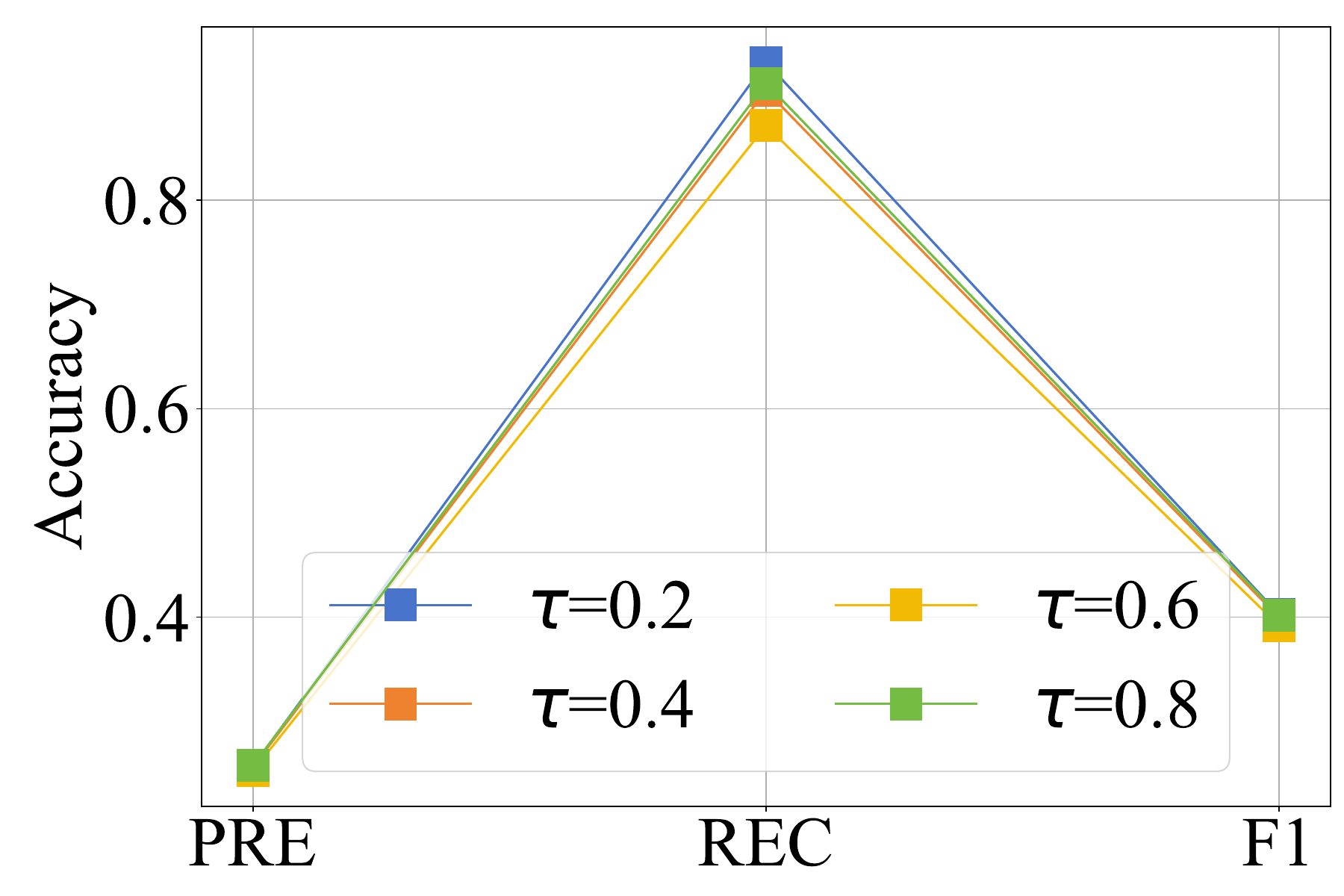}
	}

\caption{Sensitivity of the Proposed Method to Hyperparameters}
\label{env:f3}
\end{figure*}

To evaluate the impact of two key hyperparameters, the sparsity regularization strength $\lambda$ and the fusion weight $\tau$ between the long-term and short-term causal matrices, we conduct experiments on one synthetic dataset and three real-world telecommunication alarm datasets. As shown in Figure \ref{env:f3}, we test performance under varying values of $\delta$ (ranging from $0.2$ to $0.8$) and $\tau$ (ranging from $1\times10^3$ to $5\times10^3$). Overall, the method exhibits excellent performance. For example, on the synthetic dataset, PMCDTrace achieves an F1 Score as high as $0.9697$ and an SHD as low as $1$, while remaining insensitive to variations in $\lambda$.

Variations in $\lambda$ demonstrate the robustness of the proposed method. On the synthetic dataset and the 24V-439N-Microwave dataset, both F1 Score and SHD remain constant across all tested values of $\lambda$, indicating that the method is insensitive to the precise choice of regularization strength. On the other two real-world datasets, the effect of $\lambda$ is present but limited in magnitude. For example, on 18V-55N-Wireless, the F1 Score fluctuates narrowly between $0.5512$ and $0.5638$, while SHD varies between $39$ and $44$. On 25V-474N-Microwave, performance variations are similarly mild. Such strong robustness is of great practical value, as it implies that laborious hyperparameter tuning for $\lambda$ can be avoided in real-world deployments.

The variation of $\tau$ has a significant and systematic impact on causal discovery performance. On the synthetic dataset, the effect is most pronounced: as $\tau$ increases from $0.2$ to $0.8$, the F1 Score jumps from $0.52$ to $0.97$, and the SHD drops sharply from $15$ to $1$. This indicates that a higher weight assigned to short-term causality optimally balances precision and recall in this simulated scenario, which aligns with the intuition that recent effects are more influential in causal discovery. On the real-world datasets, the influence of $\tau$ is relatively stable. For instance, on 18V-55N-Wireless and 25V-474N-Microwave, the best F1 Score occurs at $\tau = 0.8$. Notably, on 24V-439N-Microwave, both F1 Score and SHD are completely insensitive to $\tau$, because events in this dataset rarely span multiple time stamps, rendering the distinction between long-term and short-term contributions negligible. These results indicate that tuning $\tau$ is crucial for optimizing performance, and that short-term weighting may be more important for causal discovery, thereby validating the effectiveness of the multi-scale fusion design.

In summary, the fusion parameter $\delta$ has a clear and important influence on performance, underscoring the necessity of integrating both short- and long-term temporal information in causal discovery. Meanwhile, the regularization strength $\lambda$ exhibits excellent robustness, indicating that the algorithmic performance does not rely on fine-tuning this constraint.

\subsection{RQ5: Ablation Experiment}

\begin{table*}[ht]
\newcommand{\tabincell}[2]{\begin{tabular}{@{}#1@{}}#2\end{tabular}}
\centering
\begin{threeparttable}
\setlength\tabcolsep{4pt}
\caption{Results of the Ablation Experiment}
\begin{tabular}{ccccccccccccccccc}
\toprule  
\multirow{2}*{Methods}& \multicolumn{4}{c}{Synthetic Datasets} & \multicolumn{4}{c}{18V-55N-Wireless} & \multicolumn{4}{c}{24V-439N-Microwave}& \multicolumn{4}{c}{25V-474N-Microwave}\\
\cline{2-17}~&PRE & REC & F1 & SHD & PRE & REC & F1 & SHD & PRE & REC & F1 & SHD & PRE & REC & F1 & SHD\\
\midrule
Baseline&	0.4545&	0.8824&	0.6000&	19&	0.4674&	0.6232&	0.5342&	55&	0.2407&	0.9489&	0.3840&	146&	0.2467&	1.0000&	0.3957&	152\\
Baseline+MSHP&	1.0000&	0.8824&	0.9375&	2&	0.5352&	0.5507&	0.5429&	51&	0.2539&	0.8394&	0.3898&	137&	0.2544&	0.8851&	0.3952&	148\\
Baseline+LSTC&	0.8421&	0.9412&	0.8889&	4&	0.4545&	0.6522	&0.5357&	56&	0.2477&	0.9927&	0.3965&	140&	0.2467&	1.0000&	0.3957&	152\\
\textbf{Proposed}&	\textbf{1.0000}&	\textbf{0.9412}&	\textbf{0.9697}&	\textbf{1}&	\textbf{0.5250}&	\textbf{0.6087}&	\textbf{0.5638}&	\textbf{39}&	\textbf{0.2604}&	\textbf{0.9124}&	\textbf{0.4052}&	\textbf{132}&	\textbf{0.2576}&	\textbf{0.9122}&	\textbf{0.4018}&	\textbf{145}\\
\bottomrule
\end{tabular}
\label{table2}
\end{threeparttable}
\end{table*}

To validate the effectiveness of the two core components in PMCDTrace, we conducted ablation experiments using S\textsuperscript{2}GCSL as the baseline. We evaluated performance on one synthetic dataset and the three real-world telecommunication alarm datasets under three configurations: adding only MSHP, adding only LSTC, and the full proposed method. As shown in Figure \ref{table2}, the full method achieves the highest F1 Score and the lowest SHD across both synthetic and all real-world datasets, demonstrating the contribution of both components to performance improvement.

On the synthetic dataset, adding MSHP alone significantly improves the F1 Score from $0.7895$ to $0.9375$, while reducing SHD from $7$ to $2$, indicating that MSHP greatly enhances precision through multi-scale modeling. Similarly, adding LSTC alone raises the F1 Score to $0.8889$ and reduces SHD to $4$, highlighting the complementary nature of the two components. On the real-world dataset 18V-55N-Wireless, adding either MSHP or LSTC leads to a modest improvement in F1 Score. Notably, MSHP contributes more to precision improvement, while LSTC contributes more to recall enhancement. On the larger-scale datasets 24V-439N-Microwave and 25V-474N-Microwave, the trends remain consistent. Adding either component alone improves the baseline performance, but combining both always yields the best overall performance, demonstrating that MSHP and LSTC are not only individually effective but also synergistic, enabling the method to adapt to real-world network data of varying scales and complexities.

In summary, the ablation experiments confirm the effectiveness of the two novel components in PMCDTrace. MSHP primarily enhances the precision of causal inference, while LSTC improves the coverage of true causal edges. The two components exhibit clear complementarity and synergy, significantly boosting the performance of the baseline method through multi-scale temporal modeling and structural regularization of the causal graph, respectively.

\section{Discussion}
\label{sec:6}
\subsection{Why does PMCDTrace work?}
The effectiveness of the PMCDTrace method stems from the synergistic collaboration of its two novel components. First, the multi-scale Hawkes process, through piecewise constant approximation, partitions historical events into different scales based on time delays, assigning independent causal matrices to each scale and fusing them. This component enables the model to finely distinguish heterogeneous dynamics in telecommunication anomaly propagation, such as instantaneous cascades and slow diffusion, fundamentally overcoming the limited expressiveness of linear Hawkes process models. Second, the long-short term constraint mechanism jointly imposes sparsity, acyclicity, and cross-scale alignment regularizations, ensuring the structural plausibility of the learned causal graph while maintaining efficient optimization.

Despite the significant improvements achieved by PMCDTrace in both causal discovery accuracy and efficiency, a limitation remains: the multi-scale partitioning relies on a pre-defined fixed time threshold. Although this partitioning method ensures computational efficiency, it may not fully adapt to the continuous and dynamic nature of anomaly propagation times in complex telecommunication networks. In future work, we plan to explore an adaptive time-scale learning mechanism that automatically partitions influence intervals through learnable threshold parameters, further enhancing the model's generalization capability across different anomaly propagation speeds.



\section{Conclusion}
\label{sec:7}
In this paper, we have addressed the challenges of insufficient multi-scale modeling and low computational efficiency in causal discovery for telecommunication network threat investigation. Specifically, we designed a multi-scale Hawkes process with a piecewise constant approximation to distinguish between short-term rapid anomaly propagation and long-term malicious infiltration. Additionally, we introduced a prior knowledge-driven long-short term constraint mechanism that jointly optimizes sparsity, acyclicity, and alignment to ensure the resulting causal graph is both plausible and robust. Extensive experimental results on six synthetic and three real-world telecommunication network alarm event datasets demonstrate that PMCDTrace significantly outperforms existing methods in both causal discovery performance and computational efficiency. In future work, we plan to explore an adaptive time-scale learning mechanism to further enhance the model's generalization capability across different anomaly propagation patterns.

\bibliography{paperbib.bib}
\bibliographystyle{IEEEtran}

\vfill
\end{document}